%% file: main.tex
\documentclass{article}

\usepackage[final,dblblindworkshop]{neurips_2026}
\workshoptitle{Interpretability as a Science}

\usepackage[utf8]{inputenc}
\usepackage[T1]{fontenc}
\usepackage{hyperref}
\hypersetup{hidelinks,
  pdftitle={How Small Can We Go? Calibrating Activation-Cache Compression for SAE Training},
  pdfauthor={Abraham Yeung}}
\usepackage{url}
\usepackage{booktabs}
\usepackage{amsfonts}
\usepackage{amsmath}
\usepackage{graphicx}
\setcitestyle{numbers,square}
\usepackage{xcolor}

\title{How Small Can We Go?\\
Calibrating Activation-Cache Compression for SAE Training}

\author{%
  Abraham Yeung\\
  Stanford University\\
  \texttt{ayeung16@stanford.edu}\\
}

\begin{document}

\maketitle
\suppressfloats[t]

\begin{abstract}
To train a sparse autoencoder (SAE), you need a massive cache of model
activations. We look into the effects that compression has on an SAE
in two ways: we keep the original fp32-trained SAE and pass the
compressed cache into the same SAE to see how many feature activations
stay the same (the frozen-SAE test), and we retrain an SAE on the
compressed cache and see how many of the original features still exist
(the retraining test). We found that in the retraining test, there is
no clear scale for retained features. On a GPT-2 cache, rounding
everything to fp16 keeps $60.6\%$ of features, but changing one value
in a million by the smallest step fp32 allows (a change eight orders
of magnitude smaller) keeps only $70.5\%$. Meanwhile, changing the
random seed keeps $23.0\%$. For the frozen-SAE test, the frozen SAE's
correlation with the original increases in order of
how precise the codecs are, but plain $L_2$ distance gives nearly the
same ordering. We also see the same frozen reading correspond to
different percentages retained at different sites. This leaves the
question of the right measurement. We build one by calibrating: retraining on
identical bytes sets the ceiling, and a
reseed sets the floor. We then measure the effect of each compressed
codec against its own site's floor. With TopK SAEs on GPT-2 small and
Pythia-1.4B at research-scale budgets, na\"ive per-token int4 loses
more features than a reseed everywhere we tested, as do PCA and random
projection on GPT-2. Standardizing each channel before quantizing
brings int4 back above the floor in terms of retained features, at the
same raw file size. At our smallest budget, a bf16 forward pass, which
some pipelines use to save memory, loses slightly more features than rotated
8-bit storage.
Calibrating a site costs four training runs and two evaluation passes.
\end{abstract}

\begin{figure}[t]
  \centering
  \includegraphics[width=\linewidth]{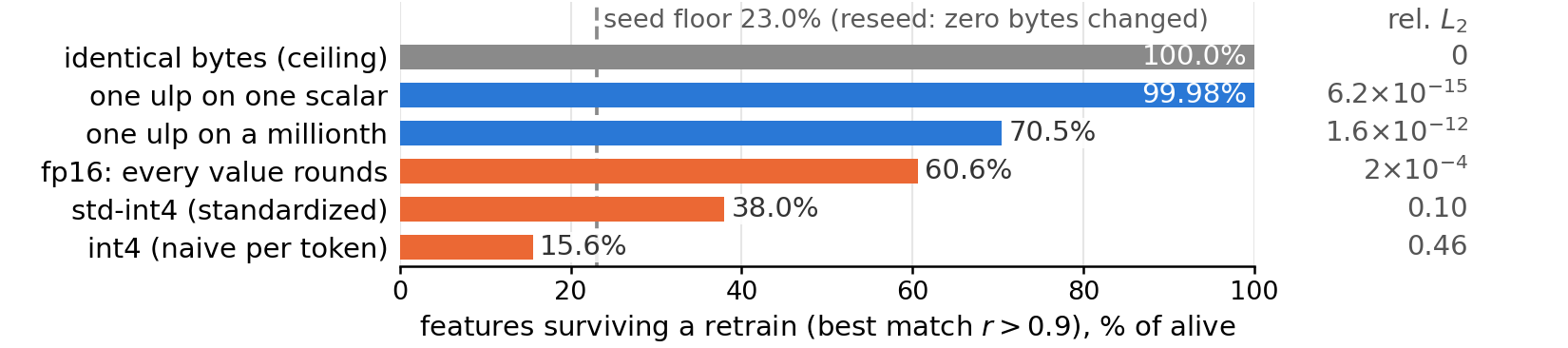}
  \caption{GPT-2 layer 6, seed 0: one ulp on a millionth of values already costs three quarters of fp16's loss, so the count
  needs a floor. Bars: alive features a retrained SAE recovers at best match $r > 0.9$ (grey:
  control; blue: tiny perturbations; orange: real codecs); dashed line: seed floor.}
  \label{fig:headline}
\end{figure}

\section{Introduction}
\label{sec:intro}

One way of understanding the internals of LLMs is to train SAEs (sparse
autoencoders) on model internal activations. However, these activations
take up a lot of space to store. A token of text is two bytes, but its
activation vector can be several kilobytes~\citep{llamascope}. This
means that training a single SAE on Gemma~2 9B can take over 1\,GiB/s
of disk read to keep the accelerators fed~\citep{gemmascope}. So, ideally we would be able
to cache the activations and compress them. SAELens, a widely used
library, stores these activations in fp32 by default~\citep{saelens}, but we
do not know what the effect of compression is on the features of the
SAE.
The closest prior work quantizes model weights rather than the
cache~\citep{duan}; separately, Paulo \& Belrose established the seed
instability we build on~\citep{paulo} (Appendix~\ref{app:related}).
An SAE learns a dictionary of sparse features, directions in
activation space that fire on some tokens; two SAEs agree on a feature
when the other SAE has a feature whose activations over the same
tokens correlate with it at $r > 0.9$.

The retraining test turns out to be very sensitive. We took a
20M-token cache and changed one number in it by the smallest amount
possible in fp32 (1 ulp, a unit in the last place). When we
retrained, $99.98\%$ of the features survived. In contrast, when we
changed a millionth of all the numbers by that same smallest step, only
$70.5\%$ of the features remained the same (Fig.~\ref{fig:headline}). Yet when we rounded every
number in the cache to fp16, a change $10^8$ times larger, we retained
$60.6\%$ of the features. So the count responds to almost any change in the bytes and,
up to fp16's distortion, only weakly to its size (\S\ref{sec:dose}).
It cannot be read
as a level, only against controls built the same way.

A measurement can be precise and still answer the wrong
question~\citep{proxypresumption,validitythreats}; the frozen-SAE test
is one, and the retraining test needs calibration first. Our two
contributions:
\begin{enumerate}
  \item \textbf{The ranking of codecs is reliable; the exact
  percentages are not} (\S\ref{sec:resolve}, \S\ref{sec:results}). The ordering holds
  across sites, thresholds and budgets (up to $2$\,pp ties, seed
  noise; TopK SAEs only), but the percentage moves with every setting
  we varied, so no single number transfers.
  \item \textbf{A same-process control lets us compare compression
  against other reasonable noise, and that is enough}
  (\S\ref{sec:protocol}, \S\ref{sec:results}). Na\"ive
  int4 alters the features more than a seed change in every whole-SAE
  comparison we ran, and per-channel standardization brings it back above
  reseeding at the same raw size. A bf16 forward pass loses $1.8$ to $3.0$\,pp more agreement than
  rotated int8 storage at our $1\times$ budget (three seeds, same
  direction, two of three outside the $2$\,pp tie band).
\end{enumerate}

\section{Calibrating the retraining test}
\label{sec:protocol}

The retraining test needs calibrating before it can be used
(\S\ref{sec:frozen} turns to the frozen-SAE test), because SAE
training is unstable across seeds~\citep{paulo,reprosubspaces}: a
na\"ive fp32-vs-codec comparison could mix the codec's effect with seed
noise. So we
calibrate against the seed change by using it as a unit of
measurement, building the bounds directly: retraining on identical
bytes gives the upper bound, and retraining with a different seed
gives the lower one.

\paragraph{Glossary of terms used in the paper.} A channel is one of
the $d$ coordinates of the activation vector. Next we have the codecs,
our compressed storage formats. rot-int$N$ standardizes each channel
(by subtracting the mean and dividing by std), applies a fixed random
rotation and quantizes to $N$ bits. std-int$N$ skips the rotation.
int$N$ is na\"ive per-token quantization (meaning one absmax scale per
token). int$N$-o$M$ keeps $M$ outlier channels in fp16. A codec's
ratio is the percentage of retained features divided by its site's
floor from reseeding. A codec that is above this reseeding floor is
considered viable. If two single-seed runs come within $2$ percentage
points (pp) we consider them tied as seed noise alone shifts every
run's number by up to $1.1$\,pp (Table~\ref{tab:m1full}). Numbers in
the text are seed 0, except for the reseeding floors which are
three-pair means and the denominator of each ratio. Tables in the
appendix add three-seed means written $\pm$ half the min--max range.
Our $1\times$ budget is 33M token-visits (tokens seen in training,
counting repeats; 24.6M on Pythia).

\paragraph{Design.} We use one fp32 master cache per site from a single
forward pass on a model. Every codec
store we make is a transcode of this cache, shard for shard. The loader then shuffles based on just the
seed, so two SAEs trained with the same seed consume identical tokens
in the same order and differ only in bytes. We trained each SAE with
Adam at learning rate $3\times10^{-4}$ for $8{,}000$ steps at batch
$4{,}096$ ($6{,}000$ on Pythia).
A site is one model, layer and read-off point; agreement
between two SAEs is the fraction of the reference SAE's alive features whose best match in the other reaches Pearson $r > 0.9$
on held-out fp32 tokens; we train
TopK SAEs (JumpReLU results are in Table~\ref{tab:jumprelu}).

\paragraph{Ceiling.} Retraining with the exact same seed and bytes
reproduces the run exactly: $r = 1.0000$, a $100\%$ agreement on Apple
M5 (MPS) and on an A100 (Appendix~\ref{app:validation}). Any same-seed
difference therefore comes from the codec alone.

\paragraph{Floor.} To find the lower bound, how much seed noise alone costs, we
retrained an SAE with a different seed. On GPT-2 small, we only retain
$23.0 \pm 0.1\%$ of features over three seed pairs at layer 6 (20M
OpenWebText tokens; TopK SAEs, where the $k{=}32$ largest of
$d_{\text{sae}}{=}12288$ latents fire per token). On Pythia-1.4B (10M
Pile tokens, bf16 forward, $d_{\text{sae}}{=}16384$), we retain
$15.7 \pm 0.1\%$ at layer 8 and $18.4 \pm 0.1\%$ at layer 16. These
floors depend on the site, so every codec must be compared against its
own site's floor.

When we reseeded only the initialization, we got an indistinguishable
$23.3 \pm 0.8\%$ (Appendix~\ref{app:validation}). Over all latents,
under a criterion like Paulo \& Belrose's, our floor is $30.6\%$, below
their GPT-2 SAEs', which differ in width, training length and
criterion~\citep{paulo}.

\paragraph{Which features we are considering.} Every number about
agreement is over alive features, those that fire on more than
$0.1\%$ of the evaluation tokens in the reference SAE, which ends up
being around $40\%$ of the dictionary at each site. Rarer features
fire too rarely for a stable correlation. Raising this cut raises agreement and the reseeding floor and lowers
rot-int4's ratio monotonically across the five cuts we swept
(Table~\ref{tab:firingrate}); ours is the loosest of the five, so a
ratio should be quoted with its cut.

\paragraph{Sanity checks.} We ran a few checks after every pipeline
change: an identity check (a cache against itself returns exactly
$1.0$), a misalignment check (one token of misalignment collapses the
correlation to $0.15$--$0.38$) and a shuffle control on best-match luck
($0.06$\,pp or less) (Appendix~\ref{app:validation}).

\section{What each test can and cannot tell us}
\label{sec:resolve}

Our two tests give very different answers for the same codec. Run the
frozen-SAE test on fp16 activations and each feature's activations
correlate with its own activations on fp32 at mean $r = 0.99997$,
every feature above $0.9$. Run the retraining test on those same fp16 bytes
and only $60.6\%$ of features are retained. Neither number can be taken at face value.

\subsection{The retraining test}
\label{sec:dose}

We first examine the effect of adding isotropic noise to the fp32
cache, at exact relative $L_2$ magnitudes from $10^{-2}$ down to
$10^{-7}$. Each level is stored in fp32 (nothing is quantized), passed
its own identity check, and got a matched SAE retrained on the
same tokens in the same order (Table~\ref{tab:dose} and
Fig.~\ref{fig:saturation}, Appendix~\ref{app:robust}). Across the four noise levels at or below
fp16's distortion, agreement barely moves: $60.6\%$ at fp16's order of distortion (by
coincidence the same number fp16 gives), rising only to $66.0\%$ three
decades below, while the frozen SAE
reads at least $0.99998$ throughout; zero perturbation returns
exactly $100\%$.

Dense noise cannot go below fp32's own precision, so for the remaining
experiments we changed only a few values instead of all of them. When
we changed $15{,}335$ scalars (a millionth of the cache's 15 billion
values) by one ulp each, $70.5\%$ of features survived. When we moved a
single scalar in the whole corpus by this amount, $99.98\%$ survived.
These two experiments differ in how many values moved and in when the
change reaches training: the single value's lasting change most likely
enters late (Appendix~\ref{app:margins}), so the $29.5$\,pp gap cannot
be put down to count alone. Most of the features
lost at $r>0.9$ have shifted rather than disappeared: lowering the
threshold to $r>0.7$ raises agreement from $70.5$ to $89.8\%$
(Table~\ref{tab:dosemargin}).

Three checks (Appendix~\ref{app:robust}) confirm that these numbers
are real. Re-running the two headline perturbations with fresh
randomness moves them by at most $1.1$\,pp. The full drop is already
present at $77$ touched values out of 15 billion ($n < 77$ remains
open), and every perturbation retains features far above the
reseeding floor.

With near losslessness in values, agreement barely responds to the size of the change
($5.4$\,pp over three decades, against $\pm1.1$\,pp of seed noise).
However, over the quantizers' range it falls from $60.6$ to
$15.6\%$ (Fig.~\ref{fig:saturation}). Any slope of agreement against
perturbation size depends on the range it is fitted over
(Table~\ref{tab:dosestrat}), so we use the retraining test only
against the reseeding floor.

\subsection{The frozen-SAE test}
\label{sec:frozen}

The frozen-SAE test asks the other question: with the same SAE, how
big is the difference in feature activations between the original and
the compressed activations? As expected, the codecs that are most precise
replicate the feature activations best. However, this is nearly
equivalent to measuring the $L_2$ distance from the original
activations.
The frozen reading orders codecs as retraining does (Spearman
$\rho = 0.982$ over 15 GPT-2 conditions, $1.000$ at both Pythia sites),
but so do the plain $L_2$ distance between caches ($0.979$) and mean
cosine ($0.982$), which need no SAE; at $n = 15$ the three are
indistinguishable (Table~\ref{tab:proxyrank}; one exception in
Appendix~\ref{app:instruments}).

But the order of the codecs does not tell us how replicated the
results are. Using the frozen-SAE test's thresholded count (features
whose frozen correlation exceeds $0.9$), the fp16 cache gets $100.0\%$
of features intact and int6 and int4-o16 keep $99.9\%$
(Table~\ref{tab:m1full}). Retraining on those same bytes, however,
returns $60.6$, $38.0$ and $37.6\%$ respectively, and the same frozen
reading of $0.9992$ corresponds to keeping $54\%$ of features at one
site and $38\%$ at another (Fig.~\ref{fig:frozen},
Table~\ref{tab:transfer}). No $L_2$ distance to the original cache can
be calibrated around the reference perturbation,
because a seed change changes no bytes, so every distance would
return $0$, yet only $23.0\%$ of features survive a reseed.

We found this pattern of viable codecs passing and na\"ive int4
failing to be not SAE-specific: linear probes and a causal steering
test (difference-in-means vectors injected into the residual stream),
each scored against its own control, got the same results with three
exceptions (PCA-256 keeps probe accuracy on one of two tasks while its
directions changed, and passes the steering test on average; int4-o16
fails one probe floor; Appendix~\ref{app:instruments}).

\section{Calibrated verdicts}
\label{sec:results}

Table~\ref{tab:m1full} has the full
thirteen-codec GPT-2 result and Fig.~\ref{fig:ratio}
(Appendix~\ref{app:budget}) plots the codecs run at every site
against each site's reseeding floor. Four conclusions follow, then the
bf16 forward.

\textbf{For viable codecs, most of the features lost would have been
lost under reseeding anyway. Na\"ive int4 loses far more.} Under
rot-int8, only $0.13\%$ of features that survive seed change
(seed-stable) are lost in all three seeds. $0.90\%$ are lost in
rot-int4 while $24.2\%$ are lost in na\"ive int4
(Table~\ref{tab:consensus}). Each viable codec retains
$88.8$--$95.7\%$ of the seed-stable features on GPT-2 and $82$--$89\%$
on Pythia. Seed-unstable features are lost to a greater degree:
$27$--$54\%$ are retained on GPT-2 and $19$--$38\%$ are retained on
Pythia. This gap stays regardless of our threshold for stability
(Table~\ref{tab:stabdef}). Reconstruction error sees none of this: the
fraction of variance unexplained (FVU) stays flat, $0.1355$ at fp16
through $0.1361$ at int4-o16, while agreement falls from $60.6$ to
$37.6\%$ (Table~\ref{tab:m1full}).

\textbf{Per-token quantization fails on outlier channels and
standardizing each channel fixes this.} A few channels have values
much larger than the rest, so one scale per token means that we waste
a lot of levels to reach the largest values. Na\"ive per-token int4
falls below the reseeding floor everywhere we tested: $15.6\%$ against
$23.0$ on GPT-2, worse at the larger model and the deeper layer,
$0.18\times$ the floor at Pythia layer 8 and $0.05\times$ at layer 16,
keeping only $13.7$ and $5.1\%$ of even the seed-stable features
(Tables~\ref{tab:m4l8}, \ref{tab:m4l16}). Standardizing each channel
before quantizing repairs it at identical raw file size: $15.6$ to
$38.0\%$ ($\pm 0.9$ over three seeds), $1.65\times$ the floor. Our
rotated codec standardizes and then applies one fixed random rotation,
so we try not rotating to separate the two effects. We find the
rotation adds nothing we can detect ($37.7 \pm 0.3$ against
$38.0 \pm 0.9\%$ over three seeds, a tie, Table~\ref{tab:ablation}).
Rotation is useful in inference quantization~\citep{quarot}, but here
standardization alone is enough (per-channel outlier handling is
standard in LLM quantization~\citep{llmint8,smoothquant,kivi}; what is
new is the measured cost of skipping it). The standardized
codecs tie or beat the outlier-channel (-o$M$) codecs at equal or
fewer bytes, with one exception the other way: int4-o16 keeps more
seed-stable features than rot-int4 ($91.2$ against $88.8\%$,
Table~\ref{tab:m1full}). With zstd on top, std-int4 stores at
$303.8$\,B/token against fp16's $1{,}419$ (Table~\ref{tab:entropy}).

\textbf{On a second model, we get the same ranking but the exact
percentages are different.} Pythia-1.4B is a model in a different
family with $11\times$ the parameters, with a different dataset and
tokenizer. We see the same codec order in terms of retained features
(the only inversions are among tied codecs) and the same 4-bit split
between standardized and na\"ive quantization (Fig.~\ref{fig:ratio}).
Na\"ive int4 performs worse than reseeding at every site and threshold
(Table~\ref{tab:threshold}).
The exact percentages, however, changed with
everything we varied (definition of stability, matching metric, budget,
sparsity, firing-rate cut, threshold). The rot codecs' ratios span
$1.06$ to $3.55\times$ (Tables~\ref{tab:sparsity}, \ref{tab:firingrate},
\ref{tab:threshold}).

\textbf{The reseeding floor rises with training, and every ratio moves
with it.} When we scale training to
$3\times$ and $10\times$, the reseeding floor rises from $23.0$ to
$42.3\%$. More optimization makes seeds agree with each other, so every
codec agreement ratio above 1 compresses toward 1 (fp16 from $2.63$ to
$1.20\times$, rot-int4 from $1.63$ to $1.09\times$). However, na\"ive
int4's agreement stays flat while the reseeding floor climbs, so it
performs worse relatively, from $0.68$ to $0.39\times$ the floor. More
training does not recover the information the codec lost. At
$3\times$, fresh text in place of our re-read cache gives the same
floor ($33.4$ against $33.3\%$) and std-int4 lead
(Appendix~\ref{app:fresh}).

\paragraph{The effect of a bf16 forward pass.} Now, we examine the
effect of something that is not a codec. If we computed the cache with
a bf16 forward pass, as Llama Scope does~\citep{llamascope},
there is no compression and yet the
frozen SAE trained on fp32 has a correlation of $r = 0.99819$. At our
$1\times$ budget, the bf16 forward costs $1.8$ to $3.0$\,pp more
retrained agreement than rotated 8-bit storage at all three seeds
($49.7$--$51.2$ against $51.6$--$53.8\%$, Table~\ref{tab:forward}; two
of the three gaps exceed the tie band), but the forward pass itself
runs $2.21\times$ faster. Now the scope is important as our budget is
33M token-visits. Gemma Scope trains each SAE on $4$--$16$B
tokens~\citep{gemmascope} and at $3\times$ our budget the gap falls
inside the tie band (one seed, Appendix~\ref{app:budget}).

\paragraph{Takeaways.} If your forward pass is bf16, store the cache
as bf16: that is bit-lossless, and the only remaining question is
whether to compress to below 16 bits. However, storing an fp32 forward
as bf16 is not free ($55.6$ against fp16's $60.6\%$,
Table~\ref{tab:m1full}). Measured against 16-bit storage rather than
fp32, rotated 4-bit is $4.0\times$ smaller before entropy coding, not
the $7.9\times$ an fp32 baseline suggests. Standardize each channel
before quantizing to 4 bits. Once that is done, 4 bits loses less
agreement than a seed change at the budgets we measured, and the
rotation on top of it adds nothing we can detect on GPT-2
(Table~\ref{tab:ablation}), so skip it.
In fp16's bytes std-int4 holds $3.95$ times the text and ties fp16 in
cross-seed agreement at $3\times$, but trains at most $1.09$ times
faster than fp32; decode on the GPU
(Appendices~\ref{app:fresh},~\ref{app:timing}).
Do not reduce dimension by PCA
or random projection ($7.8$ and $2.8\%$ on GPT-2, far below the floor,
Table~\ref{tab:m1full}), and do not certify a training codec with a
frozen SAE or with reconstruction error: the frozen SAE ranks codecs
without saying what you will keep, and reconstruction error is flat
across the viable codecs.
If you only run pre-trained SAEs over the
cache, the frozen reading is the right test (fp16, int6 and int4-o16
read $99.9$ to $100\%$ on it). Calibrating a site costs four
training runs and two evaluation passes (Appendix~\ref{app:recipe}),
and if you already plan multi-seed training~\citep{paulo} the floor's
runs are essentially free.

\section{Scope and limitations}
\label{sec:transfer}

Our compression verdicts concern activation caches for SAE training.
More broadly, any change to an SAE's training data needs
a floor and ceiling, because retraining is more sensitive than it
looks.

\paragraph{What the count measures.} Counted by what features do, the
verdict holds: removing a feature and its std-int4 match from GPT-2 shifts predictions
the same way for $38.7\%$ of features, against $23.0\%$ for a reseed.
Whether one ulp, fp16, std-int4 or a reseed caused a loss, four retrained features
rebuild only $12$--$15\%$ of the lost ones (Appendix~\ref{app:contexts}),
so a viable codec causes less of what a reseed causes. The count cannot
say whether these are the model's own features, only which ones
training finds reliably at this budget. Firing rate and activation size
in one SAE predict which of its features are seed-stable (AUC $0.85$)
and which survive a compressed cache ($0.83$--$0.84$,
Appendix~\ref{app:margins}).

\paragraph{Limitations.} Two models, residual-stream sites only,
unnormalized activations, and research-scale budgets: $10\times$ is
$328$M token-visits, and fresh text was tested at $3\times$ only. Graded
claims are TopK only: on JumpReLU the viable codecs sit within a few
points of the floor (Table~\ref{tab:jumprelu}). Pythia's reference is
itself a bf16 forward, so already perturbed. All SAEs are evaluated on
fp32 activations, and the panel omits fp8 and block-scaled formats. In
short: calibrated by its own floor and ceiling, the retraining test
finds standardized 4-bit caching viable at our budgets, na\"ive int4
not.

\clearpage
\begin{ack}
Modal provided the compute credits for the A100 experiments in
Appendix~\ref{app:timing}.
\end{ack}

\bibliographystyle{plain}

\clearpage
\appendix
\section{Full result tables}
\label{app:tables}
All appendix table cells are generated programmatically from the
released evaluation artifacts (\texttt{scripts/appendix\_tables.py}),
apart from a few constants written into the generator (such as $d = 768$). Caption prose is checked against the same
artifacts by \texttt{verify\_paper\_numbers.py}, a lint rather than a
proof.
\input{appendix_tables}
\clearpage

\section{Do matched features do the same thing?}
\label{app:contexts}
The retraining test matches features by correlation. Here we ask
whether a matched pair also does the same thing, and whether a feature
counted as lost is gone or has been cut differently. Nothing below
needs new training. The ablation and subspace analyses are scored on
the second 500k evaluation tokens, which the matching never saw. The
activating contexts and the token-type overlap use the first $150$k
and $200$k of the matching tokens. The expected outcomes of the
ablation and subspace analyses were written down before the runs
(released with the artifact).

\paragraph{Activating contexts.} Side-by-side top-activating contexts for sampled matched pairs
(fp32 vs rot-int4, and the fp32 cross-seed control) are released with the
artifact (\texttt{docs/appendix-semantic-pairs-*.md}). Illustrative
excerpts: a survived pair ($r=0.98$) fires on identical
``if you want to'' contexts on both sides with 90\% top-10
position overlap; a damaged pair ($r=0.47$) matches a precise
``rise/increase of'' feature to a coarser ``of / fall of''
feature: related, visibly degraded. The seed-pair control shows the
same qualitative signature: codec damage is seed-shaped. Bands
(survived $r>0.9$, damaged $r<0.5$) follow~\citep{duan}. Survived
pairs share $64$--$85\%$ of their top-50 activating contexts while
damaged pairs share $20$--$27\%$, and the two bands separate in all
seven comparisons we measured.

\paragraph{Ablation.} We sample 300 alive features of the fp32 SAE. For
each, we take the 32 held-out rows where it fires most strongly and
remove its own contribution (its activation times its decoder
direction) from GPT-2's residual stream at the SAE site, leaving
everything else unchanged so reconstruction error plays no part. We
record how next-token log-probabilities change where it fires, then do
the same with its best match in each retrained SAE, on the same rows.
The pair has the same effect if the two changes have cosine above $0.9$
(Table~\ref{tab:featablation}). By this causal count std-int4 keeps
$38.7\%$ of the sampled features, the same as its correlational count on
the sample, against $23.0\%$ for a reseed and $12.7\%$ for na\"ive
int4: a lead over reseeding of $15.7$\,pp (95\% CI $10.5$ to $20.8$,
bootstrap over features). Among pairs the correlation counts as matched,
std-int4's effects are as close as a reseed's (median cosine difference
$+0.014$, CI $-0.018$ to $+0.042$), so the features it keeps act like
seed-stable ones. In every condition with pairs on both sides, pairs
above $r = 0.9$ have more similar effects than pairs below $0.5$. The correlation criterion is not
sufficient on its own, though: among pairs above $r = 0.9$ the share
with the same effect runs from $94.5\%$ (one ulp on a millionth of
values) to $46.8\%$ (na\"ive int4), and is $59.0\%$ for a reseed. Per
feature the two criteria agree $79$--$86\%$ of the time (Spearman
$0.77$ between match correlation and effect cosine over $1{,}800$
pairs). The identical-bytes retrain returns cosine $1.000$; that checks
the plumbing, not the measurement's own noise, since that retrain is
bit-identical.

\paragraph{Subspaces.} Features that fail to reappear across seeds can
still span a subspace that does reappear~\citep{reprosubspaces}. If
that explained our losses, a lost feature could be rebuilt from a few
features of the retrained SAE. For each alive feature we fit a
least-squares map from its $m$ best-correlated partners' activations on
the matching tokens, and count the feature recovered if the map
explains more than $81\%$ of its held-out variance ($r > 0.9$ squared,
so $m = 1$ is the retraining count; Table~\ref{tab:subspace}). Four
partners add about $10$\,pp for a reseed ($23$ to $33\%$), and
std-int4's lead survives: $46.6$ against $33.2\%$ with four partners,
$37.9$ against $22.9\%$ with one (three runs each). Random partners
recover at most $0.12\%$. One ulp on a millionth of values is not a
change of basis: four partners take it only from $70.4$ to $74.0\%$
(three draws), well short of the $95\%$ we set in advance as the mark of
a pure re-cut. For whole sets the answer is closer to the subspace
picture: the decoder subspace of the fp32 SAE's lost features captures
the retrained SAE's unmatched features at $88$--$93\%$ of its
within-SAE value in every condition except na\"ive int4 ($73\%$); a
random subspace gets $26$--$33\%$ across conditions. The sets differ in size across
conditions, so we read that column only against its random value.

\paragraph{What this says about retraining.} Two regularities hold
whatever caused a loss: one ulp on a millionth of the values, fp16
rounding, standardized 4-bit storage or a new seed. Four partners
recover $12$--$15\%$ of the lost features in each case, and a lost
feature's best match still has a partly similar effect (median cosine
$0.41$--$0.55$ over the $8$ to $40$ pairs below $r = 0.5$ in each
condition) and fires on the same kinds of
token far more often than chance (token-type Jaccard $0.17$ for the
seed pair, against $0.007$ for random pairs). Only na\"ive int4 stands apart, recovering
$8.5\%$ and shrinking the shared subspace, though its median effect
cosine ($0.41$) is in the same range. Within this range, then, the
size of a perturbation sets how many features retraining relocates, not
what a relocated feature looks like. That is what makes a reseed a
usable unit: a codec that turns over fewer features than a reseed is
causing less of the same thing, not something different. It also limits
what surviving a reseed means. A seed-stable feature is one this
training process finds reliably at this budget, and the budget tiers
show that set growing with training (Appendix~\ref{app:budget}).
Whether sparse dictionary learning finds the features the model itself
uses is a separate question, and no count of retraining agreement, ours
included, can settle it.

\section{The protocol as a recipe}
\label{app:recipe}
To calibrate any treatment $T$ of SAE training data on a new stack:
\begin{enumerate}\setlength{\itemsep}{1pt}\setlength{\parskip}{0pt}
  \item \textbf{Fix the substrate.} One master cache per site from a
  single model forward; derive every treated dataset from it
  shard-for-shard; derive the loader's shuffle from the seed alone. Two
  runs at one seed then consume identical tokens in identical order and
  differ only in the treatment.
  \item \textbf{Measure the ceiling} (1 extra run). Retrain at the same
  seed on the same bytes. Bit-reproducible stacks return $1.0$ and any
  same-seed deviation is pure treatment effect; otherwise the measured
  ceiling is the upper bound every later number is read against.
  \item \textbf{Measure the floor} (2 extra runs). Retrain at three or
  more seeds on the untreated data and report the mean with the
  min--max range. This is the unit everything else is quoted in. Two
  further runs that reseed only the initialization bound how much the
  data order contributes; on our site the two floors are
  indistinguishable.
  \item \textbf{Run the identity and misalignment checks} (2 eval passes). Compare a
  dataset with itself through the entire pipeline (it must return
  exactly $1$), and then against itself misaligned by one token: it
  must collapse. The first decodes one store twice, so it catches nondeterminism but
  not a fault symmetric across both sides; the second is what catches a
  metric too blunt to see drift, and a pipeline that agrees regardless. Rerun both
  after every pipeline change.
  \item \textbf{Measure the saturation point} (1 extra run). Retrain on
  the untreated data perturbed by one ulp on a millionth of its scalars,
  a departure from bit-identity eight orders of magnitude below fp16.
  On our site that alone costs $29.5$\,pp of agreement. Without this
  measurement a practitioner reads fp16's $39.4$\,pp gap as codec
  damage, when a perturbation eight orders of magnitude smaller already
  costs three quarters as much. Read it as a reference point that other numbers are placed against,
  not as a term to subtract out, since drift and codec damage need not
  add: our own locality control moves it $2.6$\,pp at matched count,
  and we ran it once at one site.
  \item \textbf{Stratify, then re-stratify.} Report $T$ separately on
  features reproduced across the floor's seeds, and re-derive that split
  under a second, independent stability definition (we use weight-space
  Hungarian matching against activation-space matching). Levels move
  between definitions even when rankings do not, so quote the definition
  and threshold with every percentage.
  \item \textbf{Report the ratio as a local unit.} Absolute agreement
  does not transfer across sites, and the ratio itself moves with
  budget, sparsity and the firing-rate threshold (\S\ref{sec:results},
  Tables~\ref{tab:sparsity},~\ref{tab:firingrate});
  a number is interpretable only with all of them attached. What travels
  is the ordering within one test; across tests even
  that can reverse (\S\ref{sec:frozen}, Appendix~\ref{app:instruments}).
\end{enumerate}
Steps 2--5 are the whole cost of the calibration: four training runs and
two evaluation passes, independent of how many treatments are then
compared against them.

\section{Pipeline validation in full}
\label{app:validation}

Running a cache against itself through the whole pipeline returns
exactly $1.0$: a mean $r$ of $1.0000000$ over 500k tokens, on raw
channels at all three sites and through a frozen SAE at GPT-2 and
Pythia layer 8. We then misalign one side by a single token, and the
correlation collapses to $0.15$--$0.38$ across the sites and codecs we
ran. We rerun both controls after every change we make to the
pipeline. The identity check (our null) decodes the same store twice,
so it catches
nondeterminism but not issues that affect both sides equally.
This is why we also run the misalignment check: if our pipeline
were broken in a way that
returns high agreement no matter what, it would still pass
the identity check, but it would fail the misalignment one. On an A100
the same-seed ceiling also returns exactly $100\%$, the
floor pair reads $22.8\%$ against our $23.2$ for that pair, and rot-int4
reads $37.1$ against $37.7\%$ (\texttt{runs\_cuda\_modal/}). We also
bound one further source of inflation. Since each feature searches
$12{,}288$ candidates for its best match, some matches could just be
luck. When we shuffle the candidates, every match dies (nothing comes
near $0.9$); a stricter shuffle that preserves co-activation patterns
inflates agreement by at most $0.06$\,pp; and $96$--$99.9\%$ of matches are one-to-one for the viable codecs and
the seed pair ($90.9\%$ for na\"ive int4, $87.1\%$ for PCA-256;
released artifacts).

There are also nuances about what the floor means. Our seed impacts
both the initialization and the data order, which is a wider change
than Paulo \& Belrose's same-data floor~\citep{paulo}. When we reseeded only the
initialization and kept data order fixed, we got $23.3 \pm 0.8\%$
across three pairs, indistinguishable from the full floor. With a
weight-space metric like Paulo \& Belrose's (Hungarian matching on the
mean of encoder and decoder cosine, where they require both above
$0.7$) we get $30.6\%$ above cosine $0.7$ (over all latents, mean of
the two seed-0 pairs). Their headline figure of about the same size is for their
largest model, Llama 3 8B; their GPT-2 SAEs share roughly twice as many
latents (their Figure 5). Our setup differs from theirs in width,
matching criterion and training length, and their own sweeps show
sharing moving strongly with both width and training tokens, in
opposite directions for us (we use fewer latents and far fewer tokens),
so we cannot say which difference explains the gap.
Measuring seed instability~\citep{paulo} and choosing an
$L_0$~\citep{chanin} both require training many SAEs on the same data,
and a floor like this one is what makes such a sweep readable. The floor moves with sparsity and budget
(\S\ref{sec:results}, Table~\ref{tab:sparsity}), so it is worth measuring at the configuration
you actually use.

\paragraph{Codec details.} int$N$ quantizes each token vector
symmetrically to $N$ bits with one fp32 absolute-maximum scale per
token. int$N$-o$M$ first removes the $M$ channels with the largest mean
absolute activation on a fitting sample and stores them in fp16.
std-int$N$ standardizes each channel with a mean and standard deviation
fitted on a sample of the cache, then quantizes per token as int$N$
does; rot-int$N$ additionally applies one fixed orthogonal matrix drawn
by QR-factorizing a seeded Gaussian matrix before quantizing. Decoding
inverts every step, so the SAE always trains on fp32 reconstructions.
Pythia uses the same TopK $k{=}32$ but trains for 6{,}000 steps, so its
$1\times$ budget is 24.6M token-visits (about 2.5 passes over its 10M
tokens); correlations use 500k held-out tokens at every site.

\section{Robustness of the perturbation experiments}
\label{app:robust}

\begin{figure}[h]
  \centering
  \includegraphics[width=0.9\linewidth]{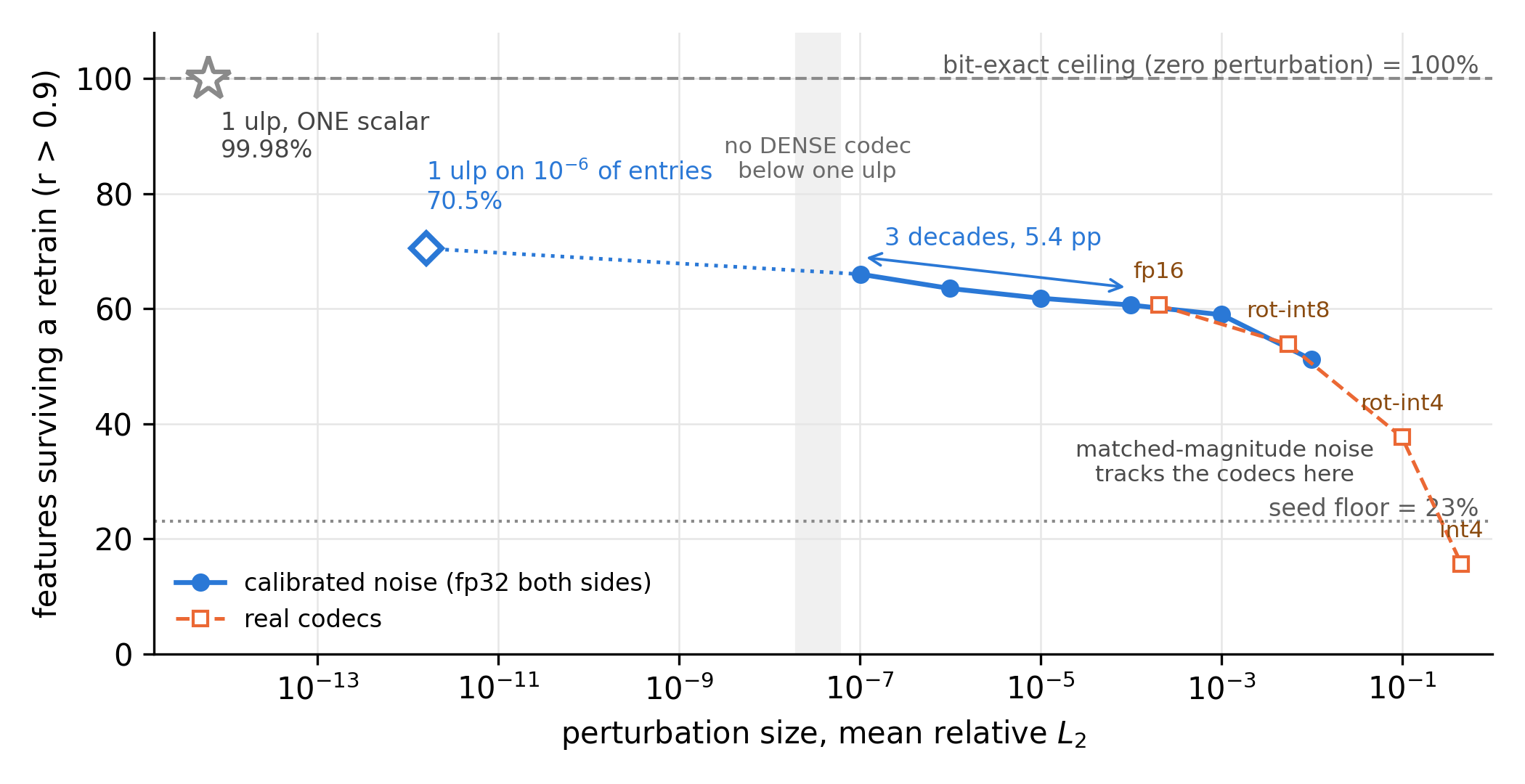}
  \caption{Retrained agreement against perturbation size. Blue:
  calibrated isotropic noise, fp32 on both sides; orange: real codecs;
  star and diamond: one ulp on one scalar and on a millionth of them;
  the shaded band marks fp32's own precision, below which no dense fp32
  noise can go. Retraining
  barely responds to the size of tiny perturbations but separates the
  real codecs clearly. GPT-2 L6, seed 0; the fp16-scale noise and the
  millionth-of-values change were each run three times and are plotted
  at their original run.}
  \label{fig:saturation}
\end{figure}

How solid are the headline numbers of \S\ref{sec:dose}? We regenerated
the noise at the fp16-scale setting and the value placements at the
millionth-of-values setting, twice each, and got $60.6 \pm 0.2$ and
$70.6 \pm 1.0\%$ across the three repetitions, so neither number is an
accident of one random draw. We also ran an experiment that packs
nearly the same count of changes into one shard of the cache (pre-registered
before it ran); it lands worse, not better ($67.9\%$, outside the
repetition spread), so where a perturbation sits matters even at fixed
count. The three experiments at $n = 77$, $153$ and $1{,}526$ changed
values land between $68.6$ and $69.9\%$, beside $70.5\%$ at
$15{,}335$: flat across two decades of count, and each was run once. The trained weights agree with these counts: the one-scalar change
moves them by $0.056$ relative Frobenius against a reseed's $1.409$
(close to $\sqrt{2}$, as expected for unaligned dictionaries; every alive latent of the one-scalar run still best-matches itself,
so no permutation enters on that side). Throughout the paper, counts at
$r>0.9$ are the number a practitioner needs; Table~\ref{tab:dose} gives
the frozen SAE's continuous reading beside them. The above/below-floor verdict of \S\ref{sec:results} also survives a
held-out-seed stratified floor: rot-int4 sits at $1.15$--$1.20\times$
that floor at all three sites and na\"ive int4 at $0.06$--$0.71\times$
(\texttt{stable\_floor.json}). A codec plus a reseed lands at the
floor for the viable codecs (fp32 seed 0 against rot-int8 and rot-int4
trained at seed 1: $24.5$ and $23.5\%$) and below it for na\"ive int4
($14.2\%$), so the viable codecs add no drift beyond a reseed. Read
against the level where retraining saturates ($66$--$70\%$) rather than
against $100$, the standardization repair is partial: std-int4 sits
about a third of the way up from the floor and rot-int8 about two
thirds.

None of this is reseeding in disguise: both of the smallest
perturbations sit far above the seed floor at every counting
threshold, by $11$--$43$\,pp for the smallest dense noise and
$12$--$47$ for the millionth-of-values change
(Table~\ref{tab:dosemargin}). Nor is it the weight-side story. In
supervised networks, one bit changed in one initial weight can
destabilize training as much as a seed change~\citep{summers}, but one
altered data scalar costs $0.02$\,pp and $15{,}335$ of them cost
$29.5$, still far short of a reseed. On the data side, timing mattered
in the one seed we tested: a $10^{-7}$ change to the first batch
alone cost about as much as the same change on every batch, and at step
$4{,}000$ almost nothing (Appendix~\ref{app:margins}; the first batch
also sets the decoder bias's starting value).

One more objection deserves its own check: seed noise averages out
across seeds, but a codec's damage might be the same features in every
seed, a shared bias that multi-seed protocols cannot average away.
Table~\ref{tab:consensus} tests this on the fp32 seed-stable set. Losses
are enriched above independence for every codec, so a shared component
exists, but under rot-int8 it amounts to $0.13\%$ of the seed-stable
features lost in all three seeds, against $24.2\%$ under na\"ive int4:
a multi-seed protocol recovers almost everything under a viable codec
and cannot under a failed one.

\section{Budget scaling in full}
\label{app:budget}

\begin{figure}[h]
  \centering
  \includegraphics[width=0.85\linewidth]{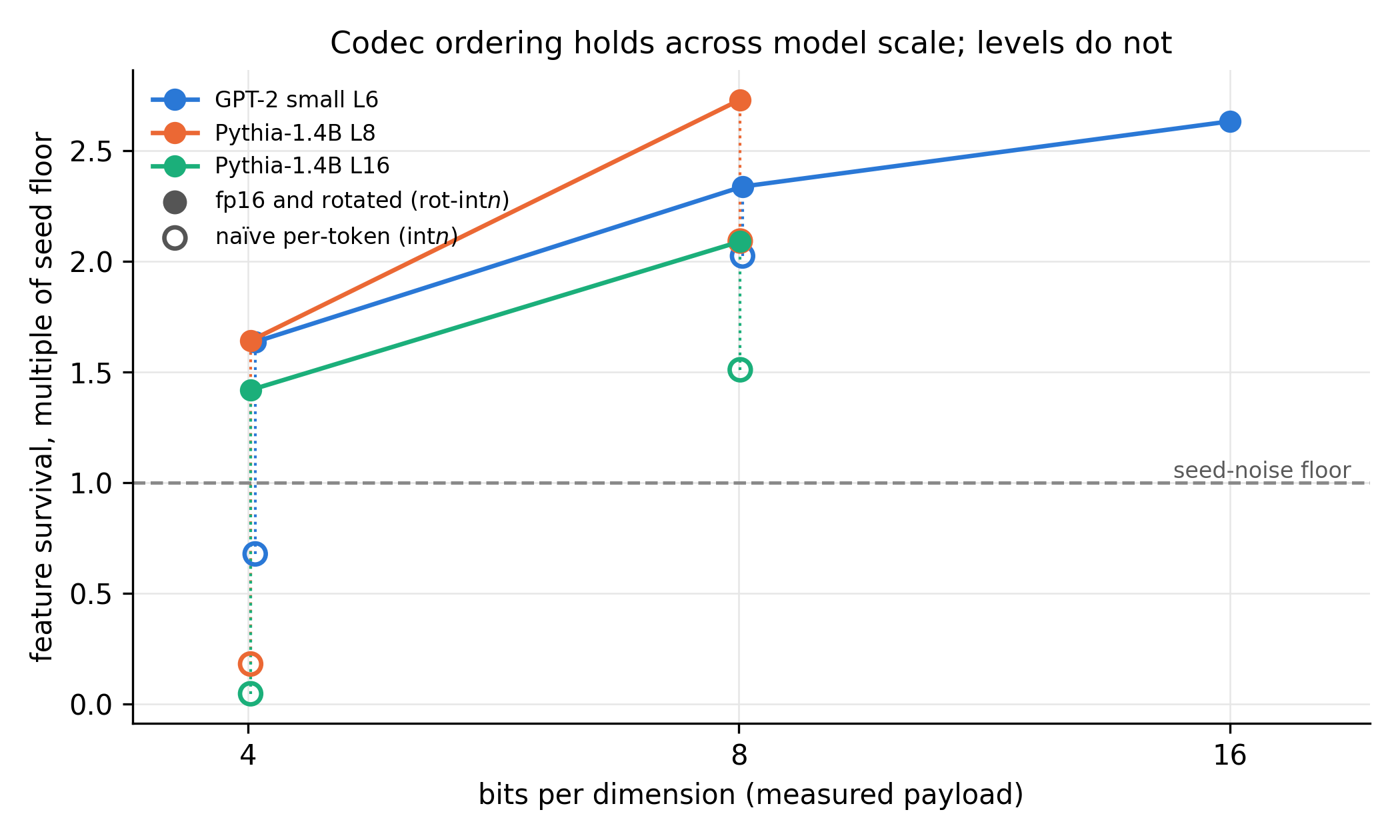}
  \caption{Agreement as a multiple of each site's own seed floor,
  against measured bits per dimension, $1\times$ budget (33M
  token-visits on GPT-2, 24.6M on Pythia). Filled markers:
  fp16 and the rotated codecs; hollow: na\"ive per-token at the same
  rate. The dotted riser is the effect of standardizing and rotating at
  identical bytes (on GPT-2, standardizing alone gives the same,
  Table~\ref{tab:ablation}), and at 4 bits it spans the floor at every
  site. Read the ordering, not the
  levels: at $10\times$ budget rot-int4's ratio is $1.09\times$.
  Pythia's fp16 is its own bf16 master, so it is omitted there.}
  \label{fig:ratio}
\end{figure}

\begin{figure}[h]
  \centering
  \includegraphics[width=0.75\linewidth]{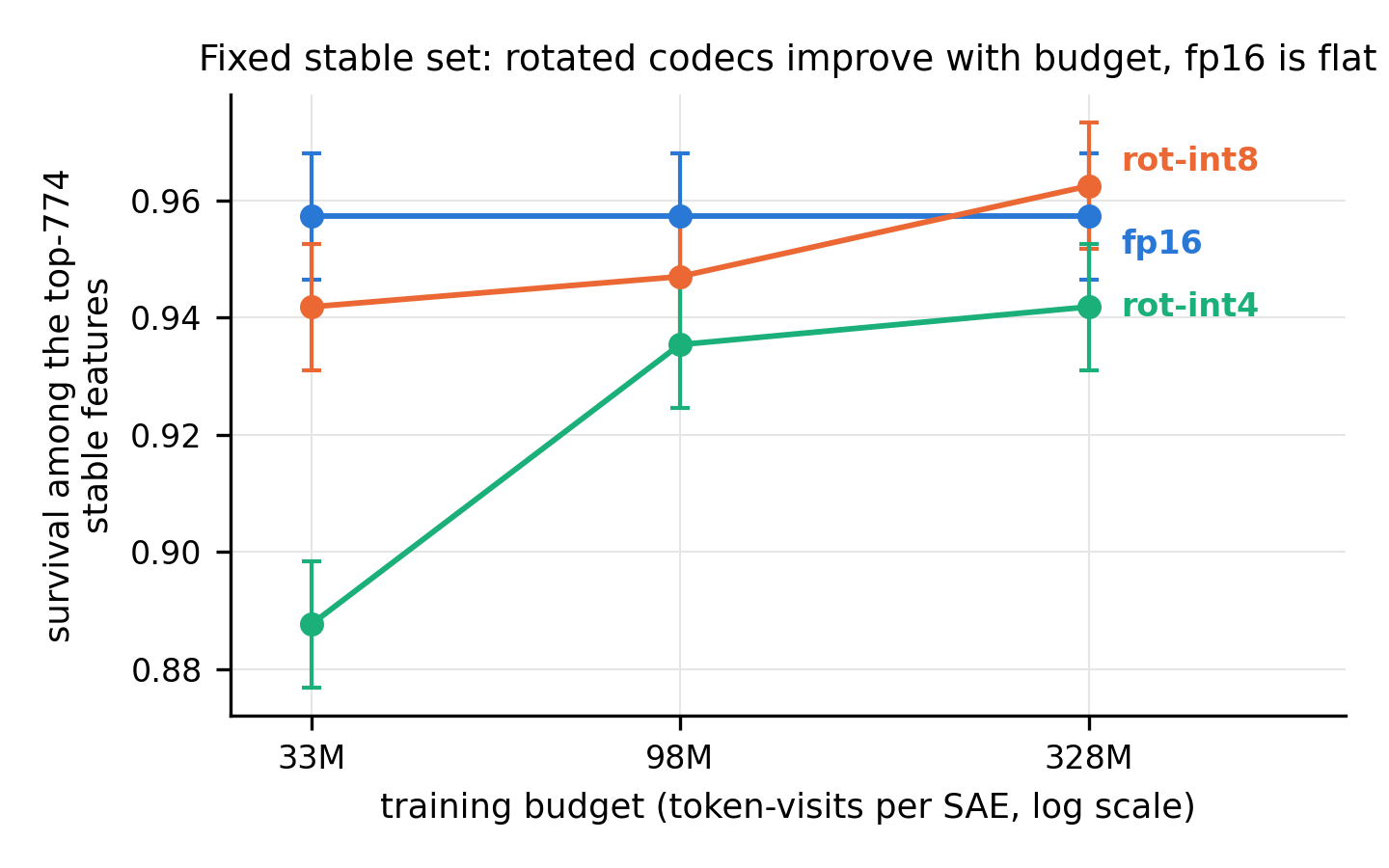}
  \caption{Agreement on a fixed-size stable set ($N = 774$, the
  $1\times$ tier's stable-feature count, taking the top $N$ by
  cross-seed stability at each budget) against training budget; error
  bars are binomial standard errors. fp16 is flat; the rotated codecs
  improve, so the gap does not widen.}
  \label{fig:budget}
\end{figure}

When we scale training to $3\times$ and $10\times$ (98M and 328M
token-visits, three fp32 seeds per tier), the floor rises from $23.0$
to $33.3$ to $42.3\%$. Every viable codec closes on the rising floor:
fp16 reads $49.2$ and $50.6\%$ at $3\times$ and $10\times$ ($1.47$ and
$1.20\times$ the floor) and rot-int4 $46.3\%$ at $10\times$
($1.09\times$).
Held at a fixed stable population, fp16 is flat while the rotated
codecs improve (Fig.~\ref{fig:budget}), so the statement we can defend
is that the 16-to-4-bit gap does not widen with budget, not that it
provably shrinks. Na\"ive int4's agreement stays flat
($15.6$/$16.3$/$16.5\%$) while the floor climbs, so it falls from
$0.68$ to $0.49$ to $0.39\times$ the floor. Table~\ref{tab:budget}
carries the per-tier numbers and two caveats: the stability threshold
rises with budget, and budget was scaled by re-reading one cache,
which isolates optimization from data diversity but not from
convergence on a fixed corpus. At $3\times$ the second
caveat no longer applies: fresh text gives the same floor and the same
std-int4 lead (Appendix~\ref{app:fresh}). At $3\times$ the bf16 forward reads $44.4$ against rot-int8's $45.3\%$
(one seed each), a $0.91$\,pp gap inside the tie band. At $1\times$ and
matched distortion that treatment is indistinguishable from a storage codec
that stores a quarter of the bytes of its fp32-stored cache
(Table~\ref{tab:forward}).

\section{Fresh text, replay and equal bytes at $3\times$}
\label{app:fresh}
Our budget tiers re-read one 20M-token cache (Appendix~\ref{app:budget}),
so the rising floor could be seeds converging on one corpus, and every
codec was compared at equal tokens rather than equal bytes. We tested
both at the $3\times$ budget (98.3M token-visits), with the outcomes
written down before any run (released with the artifact).

\paragraph{Design.} We built a 98.5M-token store whose first 77 shards
are the existing 20M-token cache and whose other 300 are new OpenWebText
documents from the same file, none shared with the training or
evaluation rows (19 rows of boilerplate that repeated an existing row
verbatim were skipped). Fresh runs read it once; replay runs read the
original cache about five times. Two checks show the two differ only in
the text: re-caching a stored shard reproduced it bit for bit, and a
fresh run's training log equals its replay twin's at all 49 logged
steps through step 4,800, diverging only once the fresh run reaches new
text (reconstruction loss $474.2$ on fresh text against $484.4$ on
replay at the next logged step). For equal bytes, std-int4 trains on the first 79.0M tokens of
the fresh store, $99.76\%$ of the fp16 cache's bytes and $3.95$ times
its tokens.

\paragraph{Replay does not inflate the floor.} The fresh floor is
$33.4\%$ against $33.3\%$ on replay, $0.09$\,pp apart
(Table~\ref{tab:fresh}), so the rise from $23.0\%$ at $1\times$ comes
from training longer, not from seeds converging on repeated text.
Same-seed fp32 SAEs trained on fresh and on replayed text still agree
on only $84.1$--$85.9\%$ of features.

\paragraph{std-int4 holds with fresh text.} It keeps $38.8\%$ against
the fresh floor's $33.4\%$ ($1.16\times$, every seed above the floor's
highest pair), as it keeps $38.7\%$ on replay ($1.16\times$; both
three-seed means). On seed-stable features the two regimes agree within
$0.17$\,pp ($88.0$ against $87.8\%$), and on the fixed 774-feature set
the budget tiers use, $93.6$ against $93.7\%$ (three seeds each).

\paragraph{Equal bytes is a tie.} Same-seed agreement would favour the
larger store, which shares the reference's data order through step
19,100 against step 4,800 for the fp16 cache, so each candidate is
scored only against fresh fp32 SAEs of the other two seeds. fp16 on 20M
tokens and std-int4 on 79.0M both reach $32.6\%$, $0.974$ of the
fresh reference's own cross-seed agreement (Table~\ref{tab:equalbytes}),
and their held-out FVU differs by only $0.46\%$. After seeing the tie
we checked, without having planned to, that this measure can separate
candidates at all: na\"ive int4 drops to $15.9\%$, while fp32 on the
20M cache lands with the others at $32.8\%$. At this budget neither the
extra text nor the extra precision moves cross-seed agreement with a
fresh-text SAE; the 4-bit cache holds $3.95$ times the text in the same
space and buys nothing measurable with it.

\section{Training time}
\label{app:timing}
Does a smaller cache train faster? On our laptop it does not: training
is compute-bound, and at $3\times$ the fp32, fp16, rot-int8 and
rot-int4 runs took $2.51$, $2.79$, $2.88$ and $2.63$ hours (a
pre-registered seven-configuration laptop grid was stopped after its
first configuration and is not reported). To test the
regime where loading matters we trained full $1\times$ runs on an A100
through Modal, with the outcome bands written down first, in seven
loader configurations that feed identical batches to identical
training: fp32, fp16 and std-int4 decoded on the CPU as in all our
other runs; std-int4 on the CPU without the identity matmul its code
path inherits from the rotated codec; and each format decoded and
shuffled on the GPU. Within a format every configuration trained
bit-identical weights, from local disk and from Modal's network volume,
on both A100 variants Modal assigned.

A smaller cache did not materially speed training
(Table~\ref{tab:timing}): the fastest std-int4 configuration ran $1.09$
times as fast as the fastest fp32 one from local disk and $1.04$ times
from the network volume, under the $1.10$ we set in advance for a
material gain. The loader mattered more than the format. With our CPU
decoder std-int4 trains $1.26$ to $1.50$ times slower than fp32 on the
same decoder ($1.28$ to $2.39$ times slower than fp32's fastest path;
our laptop loader rates are in Table~\ref{tab:throughput}), and
moving decoding and shuffling to the GPU puts every format between
$118.5$ and $129.1$ thousand tokens/s. At that rate fp32 needs about
$367$\,MB/s from storage and std-int4 about $48$\,MB/s; Modal's volume
delivered $60$ to $840$\,MB/s between runs, and the fp32 run it
starved, at $269$\,MB/s, trained at $0.41$ of the fastest fp32 run's speed. So
the case for a compressed cache is storage, $3.95$ times the text in
the bytes fp16 needs, with the same cross-seed agreement at our budget
(Appendix~\ref{app:fresh}), and speed only where storage delivers well
under what fp32 needs; if you compress, decode on the GPU. Two
caveats. The sandbox does not let us drop the page cache, so after the
copy to local disk a $7.8$\,GB std-int4 store can be served from
memory, which favours it. And Modal placed some runs on the PCIe A100;
on SXM4 runs alone the ratios are $1.08$ and $1.01$.

\section{The other tests}
\label{app:instruments}

\begin{figure}[h]
  \centering
  \includegraphics[width=0.9\linewidth]{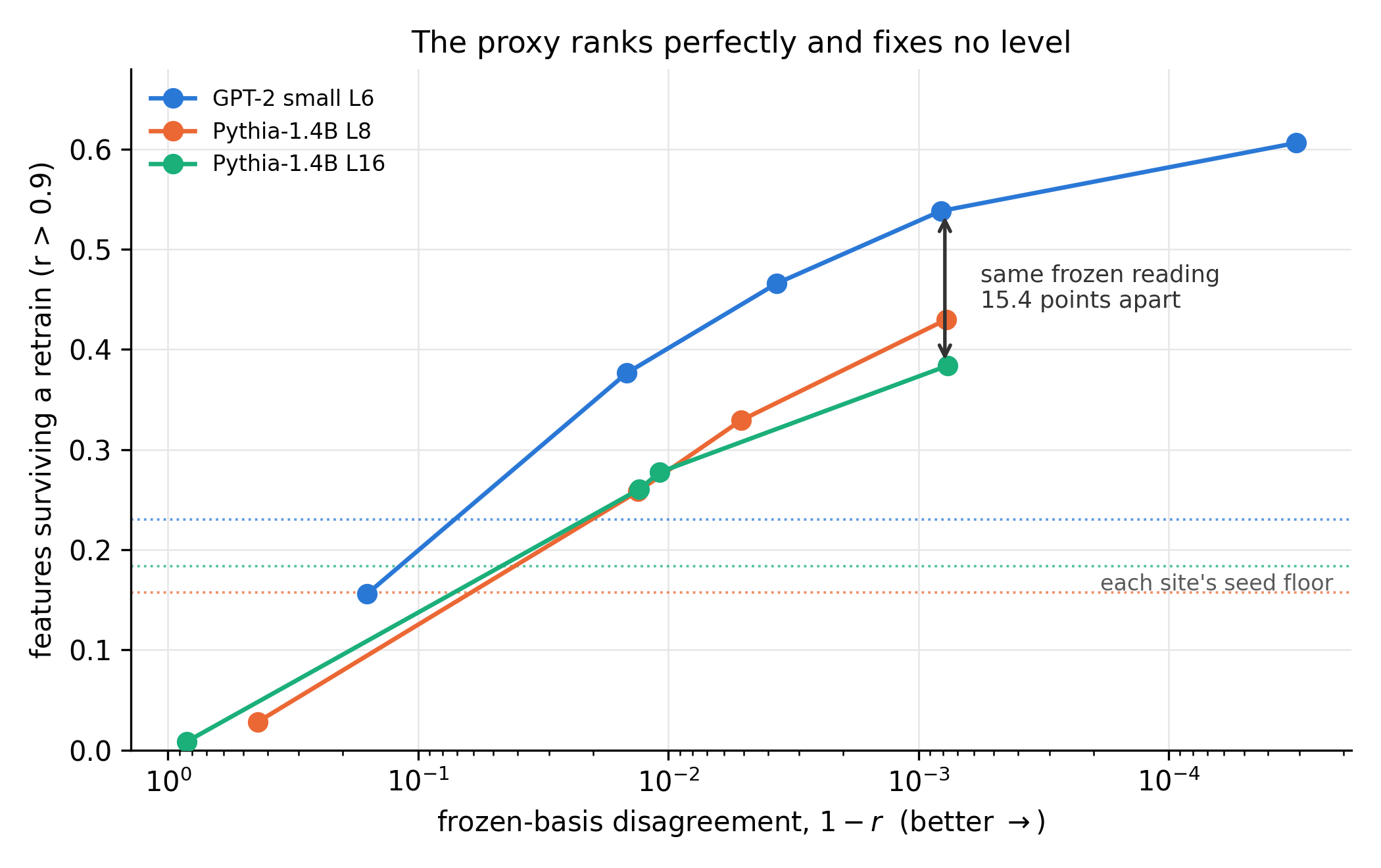}
  \caption{The frozen-SAE reading against the retraining it stands in
  for, at three sites; the $x$ axis is frozen disagreement $1-r$, log
  scale, improving rightward. Every series is monotone (the frozen SAE
  ranks codecs correctly), but the series are offset from each other
  (the same reading corresponds to different retained percentages at
  different sites); the viable codecs crowd into a span a linear axis
  would collapse to a point. Dotted lines are each site's seed floor.
  Pythia's fp16 point is omitted: the bf16 master makes it
  bit-lossless.}
  \label{fig:frozen}
\end{figure}

Linear probes, scored against their own resample floor, pass every
viable codec we probed but int4-o16 (below) and fail na\"ive int4 (Table~\ref{tab:probes}). The
exceptions cut both ways: PCA-256 keeps probe accuracy on sst2 (on ag\_news its accuracy
drops outside the floor) while its learned directions
collapse on both ($0.50$/$0.55$ cosine), which an accuracy-only check
would not show, and int4-o16 fails one probe
floor that the retraining test passes, so no single test is
uniformly the strictest. A causal steering test, scored against the
variation between two disjoint halves of the same data, fails na\"ive
int4 on both tasks and passes the three viable codecs we tested, but it also passes
PCA-256 on average, which retraining puts far below the floor
(Table~\ref{tab:steering}).
Fig.~\ref{fig:frozen} shows the cross-site picture for the frozen SAE
itself. The frozen SAE does win in one place, across codec families
and only on its continuous reading: PCA-256 (projection onto the top
256 principal components) is the smaller perturbation by both
magnitude measures yet halves na\"ive int4's agreement ($7.8$ against
$15.6\%$). Frozen mean $r$ orders that pair correctly ($0.810$ against
$0.840$; int4's value is the GPT-2 column of
Table~\ref{tab:transfer}); the thresholded reading slightly inverts it ($37.2$
against $36.7\%$, the frozen column of Table~\ref{tab:m1full}).

One selection comes close to running the other way. Among the $774$
features with the highest peak activation in the fp32 SAE (the size of
the seed-stable set), na\"ive int4 keeps $43.8\%$ against that set's
own floor of $42.1\%$ ($1.04\times$, a tie by our $2$\,pp rule; one
seed); among the $774$ with the highest mean activation it stays below
($0.69\times$; \texttt{stratify\_singlerun.json}). Every whole-SAE
comparison of na\"ive int4 in the paper is below the floor.
Which features a codec loses is partly
predictable from fp32 statistics alone (Table~\ref{tab:fragility}).

\section{Which features retraining recovers, why, and when}
\label{app:margins}
A TopK SAE is piecewise affine. The input space splits into regions
where the set of active features stays fixed, and inside each region
every active feature is an affine function of the input. The regions
meet where two pre-activations tie (or where an active one crosses
zero, a case our model below leaves out). We call a feature's margin on a
token the gap between its pre-activation and the token's runner-up,
which is the largest pre-activation left out of the top $k$. We asked
two questions about this geometry and one about training dynamics, and
wrote down how we would read each answer before running it
(\texttt{scripts/geom\_margins.py}, \texttt{scripts/geom\_identifiability.py},
\texttt{scripts/geom\_dynamics.py}).

\paragraph{What the frozen-SAE test measures.} A codec error $e$ moves
every pre-activation by $W_{\mathrm{enc}}^\top e$. A feature's frozen
activation then either shifts within its region or jumps when the token
crosses into another one: the feature drops out if it falls below the
runner-up, and a silent feature drops in if it rises past the weakest
active one. We predicted each feature's frozen disagreement $1-r$ using
only the second moment of the codec's error (estimated on held-out
tokens, with no per-token errors) and the SAE's margins on clean
activations, treating the error as Gaussian. For the 11 codecs whose
error looks like noise, the prediction ranks features with a median
Spearman $\rho$ of $0.980$ ($0.736$ for fp16, up to $0.992$), where we
had set $0.5$ in advance as the mark of an explanation. It beats the
best single feature statistic (firing rate, mean or peak activation,
median $|\rho| = 0.830$) on all 11, where we had required $9$
(Table~\ref{tab:margins}). Peak activation, the strongest single
predictor of survival when the weights are quantized~\citep{duan},
still adds a little beyond our prediction: its median absolute partial
correlation is $0.113$, just above the $0.1$ below which we had said in
advance it would count as explained. As we expected, the Gaussian model
fits worse when the codec removes part of the signal instead of adding
noise ($0.861$ for na\"ive int4, $0.877$ for PCA-256 and $0.608$ for
random projection), and for na\"ive int4 and random projection it no
longer beats the best single statistic ($0.875$ and $0.681$). For these
three, boundary crossings account for a median $84$, $46$ and $57\%$ of
a degraded or damaged feature's squared error. By the bands we set in
advance (a share of $0.5$ or more means crossings dominate, $0.3$ or
less means distortion within a region does), crossings dominate for
na\"ive int4 and random projection, and PCA-256 is mixed. Our code reproduces the frozen test's per-feature correlations
to floating-point rounding.

\paragraph{Which features retraining finds.} Results on sparse
dictionary learning tie recovery to sparse codes and incoherent
atoms~\citep{sww,agm}. That suggested two causes of seed instability:
atoms that mostly fire together (their span can be re-cut at little
cost), and atoms that point in nearly the same direction. On held-out
tokens we measured both in one SAE (co-firing with the best partner,
decoder coherence, and co-firing with the nearest decoder atom), along
with the TopK margin relative to the activation and the encoder-decoder
cosine, and compared these five with firing rate and mean and peak
activation. Seed stability turns out to be
predictable. A logistic model on all eight measures reaches a
cross-validated AUC of $0.868$ on the $1\times$, $k=32$ SAE, where
$15.9\%$ of the $4{,}860$ alive features are seed-stable, and the three
activation statistics alone reach $0.846$. Fit once there
and applied unchanged, it predicts stability in the $3\times$,
$10\times$, fresh-text, $k=16$ and $k=64$ SAEs at $0.752$ to $0.902$, all
above the $0.65$ we set in advance, as are the geometric measures alone
($0.696$ to $0.842$; Table~\ref{tab:stabpredict}). Fit to seed
stability, the models also predict which features survive retraining on
fp16, rot-int4 and std-int4 caches: $0.743$ to $0.762$ from the
geometric measures (we required $0.65$), $0.808$ to $0.828$ from all
eight, and $0.830$ to $0.841$ from the activation statistics alone. That fits \S\ref{sec:results}, where the viable
codecs lose mostly features a reseed also loses. Neither prediction
from dictionary learning held. Co-firing with a partner does not mark a
feature as unstable (AUC $0.437$, slightly the wrong way, where we
required $0.60$); nor does decoder coherence ($0.424$, also the wrong
way, a measure we set no threshold for); and the geometric measures
together do no better than the activation statistics ($0.829$ against
$0.846$, where we required a gain of $0.05$). The strongest single
predictor is the TopK margin ($0.734$). It is the strongest in five of
the six regimes, though often only just ahead of mean or peak
activation, and it tracks mean activation (Spearman $0.77$). Our
reading after the fact is that at our budgets the features training
finds reliably tend to be strongly active ones that win their TopK slot
by a wide margin, and the dictionary-learning conditions we measured do
not pick them out.

\paragraph{When features are decided.} Why does changing a millionth of
the values cost three quarters of what fp16 costs (\S\ref{sec:dose})? To find out, we
reran the seed-0 SAE step for step (its weights match
Table~\ref{tab:m1full}'s bit for bit) alongside copies whose data got
one small change and that then trained on the same batches. The change
we had planned in advance, one ulp on a single value of the first
batch, was lost to rounding and the copy stayed bit-identical. The
three readings we had planned all depended on how that copy drifted, so
none could be evaluated. We amended the plan before running the rest,
but after seeing the first $2{,}500$ steps of three dense changes at
step $0$, which grew roughly linearly and ended close together.
Of $200$ random single-value changes at step $0$, and of $200$ at step
$4{,}000$, only $58.0\%$ and $58.5\%$ change the weights after one step.
Replaying the one-scalar experiment, whose altered value enters
training at steps $112$ and $4{,}970$, reproduces that SAE bit for bit.
Its first entry changes the weights for only a few steps: they are
bit-identical to the reference at every logged step from $130$ until
the second entry (we compared weights, not the optimizer's moment
estimates). So that experiment's lasting difference most likely comes
from the change at step $4{,}970$, and its $99.98\%$ is the effect of a
late change, the reading we had written down for this case; it cannot
tell the size of a change apart from its timing. A relative change of
$10^{-7}$ to a whole batch at step $0$ ends at $64.7\%$ agreement,
against $66.0\%$ when every batch carries it, and with nearly the same
decoder distance ($0.656$ against $0.645$), inside the $5$\,pp we had
set in advance for one perturbed batch costing as much as all of them.
The same change at step $4{,}000$ ends at $99.9\%$, and from step
$6{,}000$ on at $100\%$, above the $90\%$ we had required for the change
at step $7{,}950$. Two thousand steps after the change, the copy's decoder is
$0.234$ of its norm away from the reference's when the change came at
step $0$, but only $0.045$ when it came at step $4{,}000$
(Table~\ref{tab:dynamics}). The changes at steps $0$ and $4{,}000$ end
$35$\,pp apart, which takes the label we had set in advance: damage keeps
growing with the steps remaining. Our reading after seeing the results
is sharper, that in this run features are decided early. At step $0$
the perturbed batch also sets the decoder bias's starting value, so
part of the step-$0$ cost may enter through the initialization. This is
one seed, and we did not test this change at any step between $0$ and
$4{,}000$.

\section{Extended related work}
\label{app:related}

Duan~\citep{duan} measures SAE feature survival under weight
quantization through a frozen SAE; we adopt its banding and show the
frozen protocol cannot be calibrated for the data question. Paulo \&
Belrose~\citep{paulo} and follow-ups~\citep{reprosubspaces} establish
seed instability; we turn it into the calibration a compression study
needs. MDL-SAEs~\citep{mdlsae} also frame SAEs as lossy compression
but price explanations; our object is the training corpus.
On precision itself, Gemma Scope found that switching SAE and model
weights to bf16 changes fidelity against sparsity negligibly at
inference~\citep{gemmascope}, and Llama Scope computes its activations
in bf16 to save memory~\citep{llamascope}; neither retrains on a
compressed cache against a seed floor.
Rotation-based outlier suppression is used in inference
quantization~\citep{quarot}; activation compression for
training~\citep{actnn} targets in-step memory, not a persistent
dataset. The measurement-validity
programme~\citep{proxypresumption,validitythreats,nlgmeasurement}
supplies our framing. The outlier-channel problem we hit at 4 bits, and the per-channel scaling that fixes it, are well established in LLM activation and KV-cache quantization~\citep{llmint8,smoothquant,kivi}; we do not claim the mechanism, only its measured price for SAE training.

\section{Reproduction}
\label{app:repro}
\begingroup\sloppy\emergencystretch=3em
Every number regenerates from scratch via the restartable drivers, on
one laptop except the A100 numbers of Appendices~\ref{app:validation}
and~\ref{app:timing}, which run on Modal:
\texttt{run\_m1.sh} (main grid), \texttt{run\_m1\_followup.sh}
(rotation codecs), \texttt{run\_m4.sh} (scale check),
\texttt{run\_arch.sh} (architecture panel), \texttt{run\_m6.sh}
(sparsity$\times$seed sweep), \texttt{run\_m2.sh} (probes),
\texttt{run\_controls\_a.sh} and \texttt{run\_controls\_b.sh}
(controls, including the forward-precision seeds and the four
sub-fp16 noise levels), \texttt{run\_noise\_overlap.sh} (the two noise
levels inside the codec range), \texttt{run\_sparse\_rung.sh} (the
millionth-of-values change), \texttt{run\_oneulp.sh} (the one-scalar
change), \texttt{run\_clump.sh} (the one-shard packing),
\texttt{run\_stdint4.sh} and \texttt{run\_stdint4\_seeds.sh} (the
rotation ablation), \texttt{run\_redraws.sh} (the repetitions
and the initialization-only floor), \texttt{run\_nbracket.sh} (the
count-bracket experiments),
\texttt{run\_steering.sh} (the causal steering test), \texttt{run\_budget\_scaling.sh} and
\texttt{run\_int4\_budget.sh} (budget tiers), \texttt{run\_fwdprec.sh}
and \texttt{run\_fwdprec\_b3x.sh} (the bf16 forward at $1\times$ and
$3\times$),
\texttt{run\_ksweep\_codec.sh}, \texttt{run\_pythia\_frozen.sh} and
\texttt{run\_seeds.sh}, plus \texttt{scripts/figures.py},
\texttt{scripts/appendix\_tables.py}, \texttt{scripts/budget\_fixedset.py}
and \texttt{scripts/paired\_feature\_tests.py}; Appendix~\ref{app:contexts}
reruns from \texttt{run\_subspace.sh}, \texttt{scripts/feature\_ablation.py}
and \texttt{scripts/damaged\_pair\_semantics.py}, with no training;
Appendix~\ref{app:fresh} from \texttt{run\_fresh\_b3x.sh} (with
\texttt{scripts/fresh\_store.py}), \texttt{run\_r3\_resolution.sh},
\texttt{scripts/fresh\_stable.py} and \texttt{scripts/fresh\_summary.py};
Appendix~\ref{app:timing} from \texttt{scripts/modal\_timing.py} and
\texttt{scripts/timing\_summary.py}, which deploy and read the A100 runs
on Modal;
\texttt{scripts/modal\_cuda\_ceiling.py} reruns the CUDA replication on
a cloud A100; Appendix~\ref{app:margins} reruns from
\texttt{run\_geom\_dynamics.sh} (the training dynamics),
\texttt{scripts/geom\_margins.py} and
\texttt{scripts/geom\_identifiability.py}, summarized by
\texttt{scripts/geom\_summary.py}. We use GPT-2 small~\citep{gpt2} and
Pythia-1.4B~\citep{pythia}, with text from OpenWebText~\citep{openwebtext}
and the Pile~\citep{pile}; the probe and steering tasks are
SST-2~\citep{sst2} and AG News~\citep{agnews}. Our SAEs are
TopK~\citep{topk} and JumpReLU~\citep{jumprelu}, Hungarian matching
follows~\citep{hungarian}, and the lossless entropy check uses
zstd~\citep{zstd}. A SAELens drop-in adapter
(\texttt{acc.saelens\_adapter}, zero added dependency) feeds SAELens's
trainer directly from a compressed store. Compute, summed from the recorded
per-run wall-clock in the released artifacts: 218.2 GPU-hours of
SAE training across 119 recorded runs on one Apple M5 (plus 7.0 hours for
the training-dynamics runs of Appendix~\ref{app:margins}), plus 0.47 hours
of recorded A100 SAE training for Appendix~\ref{app:validation} and 3.0
hours of A100 time for Appendix~\ref{app:timing}. Activation
caching adds 2.8 hours that are recorded (both Pythia sites); the GPT-2
cache builds, including the fresh-text store, are not in any released
manifest, and transcoding, evaluation
and probes are not separately timed. The Pythia misalignment-check values are
recorded in \texttt{runs\_m4.log} rather than in the eval JSON set, and
the frozen-SAE identity check was run at GPT-2 and Pythia layer 8
only. Peak disk was about 450\,GB.

\paragraph{Use of AI assistance.} We used an AI assistant (Claude,
Anthropic) throughout this project for code implementation, for
executing pre-registered experiment scripts, and for drafting and
editing text, including internal adversarial reviews used during
revision. No LLM is part of the method itself: the subject models
(GPT-2 and Pythia-1.4B) are the objects of measurement, and no
LLM-generated content enters any experimental result. Every number in
the paper regenerates from the released artifacts via the included
scripts, table cells are generated programmatically rather than
transcribed, \texttt{verify\_paper\_numbers.py} checks every main-text
number against those artifacts, and figures are generated by
\texttt{scripts/figures.py} from the same JSONs. All citations were
verified against the cited papers. We reviewed and take
responsibility for all content.
\endgroup

\end{document}

%% file: appendix_tables.tex
\begin{table}[htbp]
\caption{Full GPT-2 codec panel. $\pm$ rows are three-seed means ($\pm$ half the range), bare cells seed 0, so fp16 reads $59.3 \pm 1.1$ here and $60.6$ (seed 0) in the text. survived: best match above $r = 0.9$ against fp32 at the same seed; FVU: fraction of variance unexplained; mean $r$: the retrained SAE's best-match correlation, averaged over alive features (not the frozen reading); stable: survival among the $774$ features reproduced in both other fp32 seeds; Hung.\ cos: mean matched encoder and decoder cosine over all latents, dead ones included; frozen: the thresholded reading through the frozen fp32 SAE. FVU is flat across every viable codec ($0.1355$ to $0.1361$) while seed-0 agreement falls from $60.6$ to $37.6\%$. The floor row's FVU is the fp32 reference's own, and its Hungarian cosine averages the two seed-0 pairs.}
\label{tab:m1full}
\centering\footnotesize
\begin{tabular}{lrrrrrrr}
\toprule
codec & B/tok & survived \% & FVU & mean $r$ & stable \% & Hung.\ cos & frozen \% \\
\midrule
fp16 & 1536 & $59.3 \pm 1.1$ & 0.1355 & 0.866 & 95.7 & 0.817 & 100.0 \\
bf16 storage & 1536 & 55.6 & 0.1355 & 0.853 & 93.3 & 0.799 & 100.0 \\
rot-int8 & 772 & $52.7 \pm 1.1$ & 0.1356 & 0.845 & 94.2 & 0.786 & 100.0 \\
int8 & 772 & 46.6 & 0.1356 & 0.821 & 91.5 & 0.751 & 100.0 \\
int8-o8 & 780 & 50.8 & 0.1355 & 0.837 & 93.7 & 0.773 & 100.0 \\
rot-int6 & 580 & $45.9 \pm 0.7$ & 0.1355 & 0.823 & 92.1 & 0.752 & 100.0 \\
int6 & 580 & 38.0 & 0.1359 & 0.790 & 89.8 & 0.696 & 99.9 \\
int6-o8 & 590 & $44.2 \pm 0.4$ & 0.1357 & 0.813 & 91.6 & 0.740 & 100.0 \\
rot-int4 & 388 & $37.7 \pm 0.3$ & 0.1359 & 0.789 & 88.8 & 0.697 & 100.0 \\
int4-o16 & 412 & 37.6 & 0.1361 & 0.789 & 91.2 & 0.692 & 99.9 \\
int4 & 388 & $16.0 \pm 0.3$ & 0.1968 & 0.664 & 58.9 & 0.461 & 36.7 \\
pca256 & 512 & 7.8 & 0.3640 & 0.571 & 34.8 & 0.351 & 37.2 \\
rp256 & 512 & 2.8 & 0.8626 & 0.485 & 16.3 & 0.265 & 4.6 \\
\midrule
seed floor & --- & $23.0 \pm 0.1$ & 0.1355 & 0.721 & (def.) & 0.577 & --- \\
\bottomrule
\end{tabular}
\end{table}

\begin{table}[htbp]
\caption{Pythia-1.4B layer 8, same construction as Table~\ref{tab:m1full}. Codec runs are seed 0 only, so gaps under the $2$\,pp tie band are unresolved; the bracketed intervals resample features against the seed-0/seed-1 floor pair, not seeds. fp16 is bit-lossless under the bf16 master, so its row is a same-seed ceiling.}
\label{tab:m4l8}
\centering\small
\begin{tabular}{lrrr}
\toprule
codec & survived \% & $\times$ floor & stable \% \\
\midrule
fp16 (bit-identical ceiling) & 100.0 & 6.35 [6.00, 6.71] & 100.0 \\
rot-int8 & 43.0 & 2.73 [2.58, 2.87] & 88.0 \\
int8 & 32.9 & 2.09 [1.98, 2.20] & 84.9 \\
rot-int4 & 25.8 & 1.64 [1.55, 1.73] & 81.6 \\
int4 & 2.8 & 0.18 [0.15, 0.20] & 13.7 \\
\midrule
seed floor & $15.7 \pm 0.1$ & 1.00$\times$ & (def.) \\
\bottomrule
\end{tabular}
\end{table}

\begin{table}[htbp]
\caption{Pythia-1.4B layer 16. Identical construction, interval convention and bit-identical fp16 ceiling reading to Table~\ref{tab:m4l8}. Source: \texttt{runs\_m4\_L16/eval}.}
\label{tab:m4l16}
\centering\small
\begin{tabular}{lrrr}
\toprule
codec & survived \% & $\times$ floor & stable \% \\
\midrule
fp16 (bit-identical ceiling) & 100.0 & 5.44 [5.17, 5.70] & 100.0 \\
rot-int8 & 38.4 & 2.09 [1.99, 2.18] & 89.3 \\
int8 & 27.7 & 1.51 [1.44, 1.57] & 83.4 \\
rot-int4 & 26.1 & 1.42 [1.35, 1.48] & 83.0 \\
int4 & 0.8 & 0.05 [0.03, 0.06] & 5.1 \\
\midrule
seed floor & $18.4 \pm 0.1$ & 1.00$\times$ & (def.) \\
\bottomrule
\end{tabular}
\end{table}

\begin{table}[htbp]
\caption{Frozen-SAE mean $r$ beside retrained agreement at three sites. Within a site the frozen SAE ranks the codecs as retraining does; across sites the same reading carries no level (rot-int8 reads within $0.00005$ everywhere and retrains to $2.09$--$2.73\times$ the local floor). Codec rows are seed 0; floors are three-pair means.}
\label{tab:transfer}
\centering\small
\begin{tabular}{lrrrrrr}
\toprule
codec & GPT-2 $r$ & retr.\% & L8 $r$ & retr.\% & L16 $r$ & retr.\% \\
\midrule
fp16 & 0.99997 & 60.6 & 1.00000 & 100.0 & 1.00000 & 100.0 \\
rot-int8 & 0.99919 & 53.8 & 0.99922 & 43.0 & 0.99924 & 38.4 \\
int8 & 0.99632 & 46.6 & 0.99488 & 32.9 & 0.98916 & 27.7 \\
rot-int4 & 0.98536 & 37.7 & 0.98678 & 25.8 & 0.98696 & 26.1 \\
int4 & 0.84005 & 15.6 & 0.56333 & 2.8 & 0.15633 & 0.8 \\
\midrule
seed floor & --- & 23.0 & --- & 15.7 & --- & 18.4 \\
\bottomrule
\end{tabular}
\end{table}

\begin{table}[htbp]
\caption{JumpReLU panel (GPT-2, calibrated $\lambda=8.68$). The viable codecs land within a few points of the floor (fp16 $26.4$, rot-int4 $26.1$, rot-int8 $25.3$ against $23.7$ to $24.2$), too close together for a graded verdict. mean $r$: the retrained best-match correlation averaged over alive features. The fp32 and viable-codec runs end with about a third of latents dead (the na\"ive int4 run about a quarter), and the na\"ive int4 run converged to $L_0 = 38.2$ against fp32's 18.6, so its row is shown but excluded from any verdict as confounded.}
\label{tab:jumprelu}
\centering\small
\begin{tabular}{lrr}
\toprule
codec & survived \% & mean $r$ \\
\midrule
fp16 & 26.4 & 0.716 \\
rot-int8 & 25.3 & 0.708 \\
rot-int4 & 26.1 & 0.708 \\
int4 & 13.7 & 0.623 \\
\midrule
seed floor & 23.7--24.2 & 0.699 \\
\bottomrule
\end{tabular}
\end{table}

\begin{table}[htbp]
\caption{Budget scaling on a fixed stable set ($N = 774$ features at every tier, binomial SE about $1.1$\,pp). fp16 is flat, the rotated codecs improve. fp16's count is $741$ of $774$ at all three tiers by coincidence: the three pairs are distinct runs whose all-alive agreement is $60.6$, $49.2$ and $50.6\%$. Budget was scaled by re-reading one 20M-token cache (up to about 16 passes over the same tokens), so this isolates optimization from data diversity but not from convergence on a fixed corpus; at $3\times$, fresh text gives the same floor (Table~\ref{tab:fresh}).}
\label{tab:budget}
\centering\small
\begin{tabular}{lrrrrr}
\toprule
budget & stab.\ thresh & fp16 \% & rot-int8 \% & rot-int4 \% & gap pp \\
\midrule
$1\times$ (33M visits) & 0.900 & 95.7 & 94.2 & 88.8 & 7.0 \\
$3\times$ (98M) & 0.936 & 95.7 & 94.7 & 93.5 & 2.2 \\
$10\times$ (328M) & 0.954 & 95.7 & 96.3 & 94.2 & 1.6 \\
\bottomrule
\end{tabular}
\end{table}

\begin{table}[htbp]
\caption{Linear probes: accuracy change against the fp32-trained probe, and learned-direction cosine, each scored against a seven-split resample floor ($0.008$/$0.002$ accuracy and $0.80$/$0.77$ cosine on sst2/ag\_news, mean over splits, with cosine minima $0.788$/$0.759$; \texttt{probe\_floor.json}). Accuracy resolution is one validation example, so read the deltas as counts. PCA-256 keeps accuracy on sst2 but not on ag\_news (its drop there is outside the floor) while its directions collapse to about $0.5$ on both; int4-o16 fails the sst2 direction floor ($0.778$ against a floor minimum of $0.788$) on a test the retraining test passes. The two other codecs with probe runs, int8 and rot-int6, clear both floors on both tasks ($0.991$/$0.995$ and $0.998$/$0.998$ cosine) and are omitted for width.}
\label{tab:probes}
\centering\small
\begin{tabular}{lrrr}
\toprule
task & codec & $\Delta$acc & direction cos \\
\midrule
sst2 & fp16 & +0.0011 & 1.000 \\
sst2 & rot-int8 & +0.0000 & 1.000 \\
sst2 & rot-int4 & +0.0011 & 0.949 \\
sst2 & int4-o16 & -0.0080 & 0.778 \\
sst2 & int4 & -0.0608 & 0.489 \\
sst2 & pca256 & -0.0046 & 0.500 \\
ag\_news & fp16 & -0.0001 & 1.000 \\
ag\_news & rot-int8 & +0.0000 & 1.000 \\
ag\_news & rot-int4 & +0.0004 & 0.968 \\
ag\_news & int4-o16 & -0.0017 & 0.937 \\
ag\_news & int4 & -0.0554 & 0.524 \\
ag\_news & pca256 & -0.0162 & 0.547 \\
\bottomrule
\end{tabular}
\end{table}

\begin{table}[htbp]
\caption{The alive cut, swept. Tightening it raises agreement and the floor together and lowers rot-int4's ratio monotonically ($1.63$ to $1.13\times$ on GPT-2); our $0.001$ cut is the loosest of the five. rot-int4 only, so this bounds a ratio, not an ordering; the floor here averages the two seed-0 pairs at each site ($23.1$ against the three-pair $23.0$ on GPT-2).}
\label{tab:firingrate}
\centering\small
\begin{tabular}{lrrrrr}
\toprule
 & $>0.001$ & $>0.002$ & $>0.005$ & $>0.01$ & $>0.02$ \\
\midrule
\textbf{gpt2\_l6} &  &  &  &  &  \\
n alive & 4860 & 2927 & 1421 & 654 & 201 \\
floor \% & 23.1 & 27.0 & 33.7 & 40.1 & 47.5 \\
rot-int4 $\times$floor & 1.63 & 1.50 & 1.38 & 1.28 & 1.13 \\
\midrule
\textbf{pythia\_l8} &  &  &  &  &  \\
n alive & 6573 & 4053 & 1523 & 506 & 100 \\
floor \% & 15.8 & 16.1 & 20.9 & 24.7 & 36.0 \\
rot-int4 $\times$floor & 1.63 & 1.62 & 1.43 & 1.34 & 1.14 \\
\midrule
\textbf{pythia\_l16} &  &  &  &  &  \\
n alive & 7088 & 4510 & 1525 & 353 & 73 \\
floor \% & 18.5 & 19.2 & 25.7 & 46.2 & 58.2 \\
rot-int4 $\times$floor & 1.41 & 1.37 & 1.30 & 1.22 & 1.06 \\
\bottomrule
\end{tabular}
\end{table}

\begin{table}[htbp]
\caption{Sparsity moves the floor without touching a byte: over a $1.95\times$ swing in the floor, rot-int8's ratio spreads $1.72$--$3.23\times$ while its survival among seed-stable features spreads only $92.8$--$97.1\%$.}
\label{tab:sparsity}
\centering\small
\begin{tabular}{lrrrrrr}
\toprule
sparsity & seed floor \% & rot-int8 \% & $\times$floor & stable-only \% & $n_{\text{alive}}$ & stable frac \% \\
\midrule
$k=16$ & $28.6 \pm 0.8$ & 49.1 & 1.72$\times$ & 92.8 & 2,912 & 21.0 \\
$k=32$ & $23.0 \pm 0.1$ & 53.8 & 2.34$\times$ & 94.2 & 4,860 & 15.9 \\
$k=64$ & $14.7 \pm 0.2$ & 47.4 & 3.23$\times$ & 97.1 & 8,269 & 10.1 \\
\bottomrule
\end{tabular}
\end{table}

\begin{table}[htbp]
\caption{A treatment that is not a storage codec: computing the cache with a bf16 forward (stored in fp32) beside the storage codecs. Over three seeds the bf16 forward retains $49.7$--$51.2\%$ against rot-int8's $51.6$--$53.8$, disjoint but with $n=3$ (a sign test cannot go below $p=0.25$; within-seed deltas $-2.7$, $-1.8$, $-3.0$\,pp). At $3\times$ budget the comparison reads $44.4$ against $45.3\%$ (one seed each), inside the tie band. It runs $2.21\times$ faster.}
\label{tab:forward}
\centering\footnotesize
\begin{tabular}{lrrrrrrrr}
\toprule
treatment & B/tok & rel-L2 & frozen $r$ & survived \% & $\times$floor & stable \% & Hung.\ cos & FVU \\
\midrule
fp16 storage & 1536 & 0.00020 & 0.99997 & 60.6 & 2.63 & 95.7 & 0.817 & 0.1355 \\
rot-int8 storage & 772 & 0.00551 & 0.99919 & 53.8 & 2.34 & 94.2 & 0.786 & 0.1356 \\
int8-o8 storage & 780 & 0.00928 & 0.99868 & 50.8 & 2.21 & 93.7 & 0.773 & 0.1355 \\
\textbf{bf16 forward} & \textbf{3072} & \textbf{0.00980} & \textbf{0.99819} & \textbf{51.2} & \textbf{2.22} & \textbf{94.2} & \textbf{0.775} & \textbf{0.1356} \\
int8 storage & 772 & 0.02660 & 0.99632 & 46.6 & 2.02 & 91.5 & 0.751 & 0.1356 \\
\midrule
\textbf{seed floor} & --- & --- & --- & \textbf{23.0} & \textbf{1.00} & (def.) & --- & --- \\
\midrule
\textbf{bf16 fwd, 3 seeds} & --- & --- & --- & \textbf{49.7--51.2} & --- & --- & --- & --- \\
rot-int8, 3 seeds & --- & --- & --- & 51.6--53.8 & --- & --- & --- & --- \\
\bottomrule
\end{tabular}
\end{table}

\begin{table}[htbp]
\caption{Lossless zstd-3 over the quantized payloads, measured on one shard per codec. Rotated codecs compress only $1.04$--$1.26\times$ further; na\"ive scaling leaves headroom that zstd recovers. Standardization without the rotation lands with the rotated codec ($303.8$ against $307.5$\,B/token).}
\label{tab:entropy}
\centering\small
\begin{tabular}{lrrr}
\toprule
codec & raw B/tok & zstd B/tok & ratio \\
\midrule
fp16 & 1536 & 1419.0 & 1.08$\times$ \\
int8 & 772 & 510.9 & 1.51$\times$ \\
rot-int8 & 772 & 709.1 & 1.09$\times$ \\
int6 & 580 & 431.9 & 1.34$\times$ \\
rot-int6 & 580 & 557.0 & 1.04$\times$ \\
int4 & 388 & 146.4 & 2.65$\times$ \\
rot-int4 & 388 & 307.5 & 1.26$\times$ \\
std-int4 & 388 & 303.8 & 1.28$\times$ \\
int4-o16 & 412 & 309.1 & 1.33$\times$ \\
\bottomrule
\end{tabular}
\end{table}

\begin{table}[htbp]
\caption{Serial loader throughput per store (prefetch $= 0$, warm page cache, one Apple M5). The $1\times$ training run consumes $10{,}017$ tokens/s, so even the slowest store (int4-o16, 55k tokens/s) outruns training; the standardize-and-rotate decode path costs real time (rot-int8 reads at $64\%$ of int8's rate). One-machine numbers. In a paired idle-machine measurement (two interleaved runs each), std-int4 decodes at 88\,MB/s against rot-int4's 89; our std-int4 keeps the rotation matmul with an identity matrix, so this does not isolate the rotation's own cost.}
\label{tab:throughput}
\centering\small
\begin{tabular}{lrrrr}
\toprule
codec & B/tok read & disk MB/s & decode MB/s & loader tok/s \\
\midrule
fp32 & 3072 & 3094 & 10416 & 383k \\
fp16 & 1536 & 3085 & 4748 & 445k \\
int8 & 772 & 2743 & 754 & 319k \\
rot-int8 & 772 & 2917 & 242 & 205k \\
int6 & 580 & 2796 & 208 & 228k \\
rot-int6 & 580 & 2998 & 113 & 147k \\
int4 & 388 & 2677 & 185 & 270k \\
rot-int4 & 388 & 2543 & 86 & 162k \\
int4-o16 & 412 & 3037 & 25 & 55k \\
pca256 & 512 & 2603 & 402 & 335k \\
\bottomrule
\end{tabular}
\end{table}

\begin{table}[htbp]
\caption{Ratio to floor as the counting threshold moves. Levels change with the threshold; the ordering and the below-floor verdict for na\"ive int4 do not.}
\label{tab:threshold}
\centering\small
\begin{tabular}{lrrrrr}
\toprule
 & $r>0.7$ & $r>0.8$ & $r>0.85$ & $r>0.9$ & $r>0.95$ \\
\midrule
\textbf{GPT-2 L6} &  &  &  &  &  \\
floor (survival) & 59.4 & 42.4 & 32.9 & 23.0 & 10.9 \\
fp16 & 1.45$\times$ & 1.83$\times$ & 2.15$\times$ & 2.63$\times$ & 3.78$\times$ \\
rot-int8 & 1.39$\times$ & 1.71$\times$ & 1.97$\times$ & 2.34$\times$ & 3.19$\times$ \\
rot-int4 & 1.22$\times$ & 1.39$\times$ & 1.50$\times$ & 1.63$\times$ & 1.90$\times$ \\
int4 & 0.83$\times$ & 0.79$\times$ & 0.75$\times$ & 0.68$\times$ & 0.61$\times$ \\
\midrule
\textbf{Pythia L8} &  &  &  &  &  \\
floor (survival) & 56.3 & 36.5 & 26.3 & 15.7 & 6.0 \\
fp16 & 1.78$\times$ & 2.74$\times$ & 3.80$\times$ & 6.35$\times$ & 16.80$\times$ \\
rot-int8 & 1.43$\times$ & 1.86$\times$ & 2.19$\times$ & 2.73$\times$ & 3.55$\times$ \\
rot-int4 & 1.20$\times$ & 1.37$\times$ & 1.49$\times$ & 1.64$\times$ & 1.72$\times$ \\
int4 & 0.43$\times$ & 0.33$\times$ & 0.25$\times$ & 0.18$\times$ & 0.07$\times$ \\
\midrule
\textbf{Pythia L16} &  &  &  &  &  \\
floor (survival) & 56.9 & 38.9 & 28.9 & 18.4 & 7.7 \\
fp16 & 1.76$\times$ & 2.57$\times$ & 3.46$\times$ & 5.44$\times$ & 12.98$\times$ \\
rot-int8 & 1.34$\times$ & 1.60$\times$ & 1.81$\times$ & 2.09$\times$ & 2.50$\times$ \\
rot-int4 & 1.14$\times$ & 1.25$\times$ & 1.36$\times$ & 1.42$\times$ & 1.55$\times$ \\
int4 & 0.17$\times$ & 0.10$\times$ & 0.08$\times$ & 0.05$\times$ & 0.02$\times$ \\
\bottomrule
\end{tabular}
\par\smallskip\footnotesize Rows are a four-codec slice of a sweep covering 13 codecs at GPT-2 and these four at each Pythia site. Ordering is invariant across it at both Pythia sites. On GPT-2 the only inversions are among rot-int4, int4-o16 and int6 (the last two not shown here), whose largest inverting gap is 0.45\,pp, inside the tie band. Na\"ive int4 is below floor at every threshold at every site.
\end{table}

\begin{table}[htbp]
\caption{The noise levels and sparse changes (seed 0; pp/decade is the change in agreement per decade of relative $L_2$ to the next row): isotropic noise at an exact relative $L_2$ and one-ulp changes to a fixed count of values, all stored in fp32 on both sides. Everything from $n=77$ to $15{,}335$ is flat; the one-value change also differs in when it reaches training (Appendix~\ref{app:margins}). The fp16-scale noise and the millionth-of-values change were re-run twice with fresh randomness ($60.6$/$60.4$/$60.8$ and $70.5$/$69.7$/$71.6\%$); the table prints the original runs.}
\label{tab:dose}
\centering\small
\begin{tabular}{lrrrr}
\toprule
perturbation & rel-L2 & frozen $r$ & survived \% & pp/decade \\
\midrule
zero (bit-exact) & 0 & 1.000000 & \textbf{100.0} & --- \\
noise $10^{-2}$ & 1.00e-02 & 0.998551 & 51.2 & +7.82 \\
noise $10^{-3}$ & 1.00e-03 & 0.999847 & 59.0 & +2.38 \\
fp16 (codec) & 2.03e-04 & 0.999969 & 60.6 & +0.07 \\
noise $10^{-4}$ & 1.00e-04 & 0.999985 & 60.6 & +1.19 \\
noise $10^{-5}$ & 1.00e-05 & 0.999999 & 61.8 & +1.69 \\
noise $10^{-6}$ & 1.00e-06 & 1.000000 & 63.5 & +2.50 \\
noise $10^{-7}$ & 1.01e-07 & 1.000000 & 66.0 & --- \\
1 ulp on $10^{-6}$ of entries ($\sim$15k) & 1.59e-12 & 1.000000 & 70.5 & --- \\
1 ulp on 1{,}526 scalars & 1.70e-13 & 1.000000 & 68.6 & --- \\
1 ulp on 153 scalars & 9.57e-15 & 1.000000 & 69.3 & --- \\
1 ulp on 77 scalars & 5.61e-15 & 1.000000 & 69.9 & --- \\
1 ulp on \textbf{one} scalar & 6.24e-15 & 1.000000 & 99.98 & --- \\
\midrule
seed floor & --- & --- & 23.0 & --- \\
\bottomrule
\end{tabular}
\end{table}

\begin{table}[htbp]
\caption{The noise levels and codecs on the seed-stable stratum ($n = 774$) beside all alive features. The response is flatter still on the stable stratum ($0.35$\,pp per decade from noise $10^{-7}$ to $10^{-4}$, against $2.6$ from fp16 to rot-int4), and the ratio of the two rates depends on the window, so we quote no single ratio.}
\label{tab:dosestrat}
\centering\small
\begin{tabular}{lrrr}
\toprule
arm & rel-L2 & all alive \% & seed-stable \% \\
\midrule
noise $10^{-7}$ & 1.01e-07 & 66.0 & 96.4 \\
noise $10^{-6}$ & 1.00e-06 & 63.5 & 97.0 \\
noise $10^{-5}$ & 1.00e-05 & 61.8 & 96.1 \\
noise $10^{-4}$ & 1.00e-04 & 60.6 & 95.3 \\
noise $10^{-3}$ & 1.00e-03 & 59.0 & 95.9 \\
noise $10^{-2}$ & 1.00e-02 & 51.2 & 93.4 \\
fp16 & 2.03e-04 & 60.6 & 95.7 \\
rot-int8 & 5.51e-03 & 53.8 & 94.2 \\
rot-int4 & 9.99e-02 & 37.7 & 88.8 \\
na\"ive int4 & 4.55e-01 & 15.6 & 58.9 \\
\bottomrule
\end{tabular}
\end{table}

\begin{table}[htbp]
\caption{How far the two smallest perturbations sit above the seed floor at each counting threshold. The margin is positive everywhere ($11$--$43$\,pp dense, $12$--$47$ sparse) but its size is threshold-dependent.}
\label{tab:dosemargin}
\centering\small
\begin{tabular}{lrrrr}
\toprule
criterion & perturbed \% & floor \% & margin (pp) & below ceiling (pp) \\
\midrule
\textbf{$10^{-7}$ dense} &  &  &  &  \\
$r>0.50$ & 95.6 & 83.8 & +11.8 & 4.4 \\
$r>0.70$ & 88.3 & 59.4 & +28.9 & 11.7 \\
$r>0.85$ & 76.2 & 32.9 & +43.3 & 23.8 \\
$r>0.90$ (ours) & 66.0 & 23.0 & +43.0 & 34.0 \\
$r>0.95$ & 46.3 & 10.9 & +35.4 & 53.7 \\
$r>0.99$ & 12.2 & 0.8 & +11.4 & 87.8 \\
\midrule
\textbf{sparse, 1 ulp on $10^{-6}$} &  &  &  &  \\
$r>0.50$ & 96.1 & 83.8 & +12.3 & 3.9 \\
$r>0.70$ & 89.8 & 59.4 & +30.4 & 10.2 \\
$r>0.85$ & 79.3 & 32.9 & +46.4 & 20.7 \\
$r>0.90$ (ours) & 70.5 & 23.0 & +47.4 & 29.5 \\
$r>0.95$ & 52.1 & 10.9 & +41.3 & 47.9 \\
$r>0.99$ & 15.1 & 0.8 & +14.2 & 84.9 \\
\bottomrule
\end{tabular}
\end{table}

\begin{table}[htbp]
\caption{The causal test: a difference-in-means steering vector injected into the residual stream, scored against the variation between two disjoint halves of the same data (the half-versus-half floor runs about twice a full-data comparison, so it is conservative only for a fail; the like-for-like referent is half the floor). Na\"ive int4 fails at least $399$ of $400$ resample splits on both tasks; PCA-256 passes on average but fails $51$ of $400$ ag\_news splits against the like-for-like referent; task-matched prompts and a strength sweep leave every verdict unchanged.}
\label{tab:steering}
\centering\small
\begin{tabular}{lrrr}
\toprule
codec & direction cos & causal cos & effect/floor \\
\midrule
\textbf{ag\_news} (floor 0.9995 / 0.9998) &  &  &  \\
fp16 & 1.0000 & 1.0000 & 0.002 \\
rot-int8 & 1.0000 & 1.0000 & 0.007 \\
rot-int4 & 1.0000 & 1.0000 & 0.053 \\
pca256 & 0.9998 & 0.9999 & 0.276 \\
int4 & 0.7901 & 0.8294 & 13.713  \textbf{FAIL} \\
\midrule
\textbf{sst2} (floor 0.9851 / 0.9877) &  &  &  \\
fp16 & 1.0000 & 1.0000 & 0.001 \\
rot-int8 & 1.0000 & 1.0000 & 0.003 \\
rot-int4 & 0.9999 & 1.0000 & 0.056 \\
pca256 & 0.9991 & 0.9998 & 0.245 \\
int4 & 0.8763 & 0.9702 & 2.354  \textbf{FAIL} \\
\bottomrule
\end{tabular}
\end{table}

\begin{table}[htbp]
\caption{Which step repairs 4-bit. std-int4 is byte-identical to rot-int4 with the rotation set to the identity, so the two differ by exactly the rotation: standardization alone gives $38.0 \pm 0.9$\%, adding the rotation $37.7 \pm 0.3$\%, inside the tie band. All three arms are three-seed means over same-seed pairs. The ablation keeps the rotation matmul, so it isolates the rotation's effect, not its decode cost.}
\label{tab:ablation}
\centering\small
\begin{tabular}{lrrrrr}
\toprule
arm & B/tok & rel-$L_2$ & frozen $r$ & survived \% & $\times$floor \\
\midrule
na\"ive int4 & 388 & 0.4553 & 0.84005 & $16.0 \pm 0.3$ & 0.69$\times$ \\
\textbf{std-int4} (standardize only) & 388 & 0.1033 & 0.98466 & $38.0 \pm 0.9$ & 1.65$\times$ \\
rot-int4 (standardize $+$ rotate) & 388 & 0.0999 & 0.98536 & $37.7 \pm 0.3$ & 1.64$\times$ \\
\bottomrule
\end{tabular}
\end{table}

\begin{table}[htbp]
\caption{Is codec damage a shared bias or seed-like noise? On the fp32 seed-stable set ($n=774$), losses are enriched above independence for every codec, so a shared component exists, but under rot-int8 only $0.13\%$ of seed-stable features are lost in all three seeds against $24.2\%$ under na\"ive int4. Recomputed from released pair files; no training.}
\label{tab:consensus}
\centering\small
\begin{tabular}{lrrrrr}
\toprule
codec & lost/seed \% & P(lost s1 $|$ lost s0) \% & indep.\ \% & lost all 3 \% & indep.\ \% \\
\midrule
rot-int8 & 6.1 & 17.8 & 6.3 & 0.13 & 0.023 \\
rot-int4 & 11.1 & 25.3 & 12.0 & 0.90 & 0.134 \\
int4 & 41.0 & 75.5 & 42.9 & 24.16 & 6.876 \\
\bottomrule
\end{tabular}
\end{table}

\begin{table}[htbp]
\caption{Survival among seed-stable / seed-unstable features under four definitions of stability: activation-space matching at three thresholds, and weight-space Hungarian matching on encoder$+$decoder cosine. Under every definition the viable codecs keep far more of the stable features than of the unstable ones (stable 62.9 to 98.8\%, unstable 19.9 to 57.9\%), and na\"ive int4 is lowest on both; the levels move with the definition, so a stratified percentage is only interpretable with its definition attached. On Pythia the viable codecs keep 18.6 to 38.3\% of the unstable features (\texttt{runs\_m4\_L8}, \texttt{runs\_m4\_L16} stability files). GPT-2 L6, seed 0. Sources: \texttt{stability\_fp32\_s0\_t0p*.json}, \texttt{stability\_hungarian.json}.}
\label{tab:stabdef}
\centering\small
\begin{tabular}{lrrrr}
\toprule
codec & $r>0.8$ \% & $r>0.9$ \% & $r>0.95$ \% & Hungarian \% \\
\midrule
fp16 & 90.1 / 46.1 & 95.7 / 54.0 & 97.3 / 57.9 & 80.7 / 52.7 \\
rot-int8 & 86.0 / 38.0 & 94.2 / 46.2 & 98.8 / 50.5 & 76.4 / 44.9 \\
rot-int4 & 73.8 / 19.9 & 88.8 / 28.0 & 95.5 / 33.3 & 62.9 / 27.7 \\
int4 & 41.3 / 2.9 & 58.9 / 7.4 & 73.6 / 11.3 & 26.9 / 11.1 \\
\bottomrule
\end{tabular}
\end{table}

\begin{table}[htbp]
\caption{Spearman rank agreement of three cheap readings with retrained agreement over every released condition per site. The 15 GPT-2 conditions are the 13 storage codecs of Table~\ref{tab:m1full} plus the bf16 forward and the std-int4 ablation. None of the three needs a retrained SAE, and none ranks better than the others.}
\label{tab:proxyrank}
\centering\small
\begin{tabular}{lrrrr}
\toprule
site & $n$ & frozen $r$ $\rho$ & rel-$L_2$ $\rho$ & cosine $\rho$ \\
\midrule
GPT-2 L6 & 15 & 0.982 & 0.979 & 0.982 \\
Pythia L8 & 5 & 1.000 & 1.000 & 1.000 \\
Pythia L16 & 5 & 1.000 & 1.000 & 1.000 \\
\bottomrule
\end{tabular}
\end{table}

\begin{table}[htbp]
\caption{Predicting per-feature survival from fp32 statistics. CV AUC: a logistic model on all three, 5-fold cross-validated; the other columns are single-statistic AUCs. Mean activation is the strongest single predictor, unlike the peak activation reported for weight quantization.}
\label{tab:fragility}
\centering\small
\begin{tabular}{lrrrr}
\toprule
codec & CV AUC & mean act & peak act & rate \\
\midrule
rot-int6 & 0.836 & 0.758 & 0.713 & 0.552 \\
rot-int4 & 0.844 & 0.755 & 0.711 & 0.570 \\
int4-o16 & 0.831 & 0.749 & 0.711 & 0.564 \\
int6 & 0.838 & 0.747 & 0.709 & 0.573 \\
\bottomrule
\end{tabular}
\end{table}

\begin{table}[htbp]
\caption{Do matched features do the same thing? 300 sampled alive fp32 features, each removed from the residual stream on its 32 strongest held-out rows, then its best match in each retrained SAE on the same rows. Same effect: cosine above $0.9$ between the two changes in next-token log-probabilities; the last three columns are medians. std-int4 exceeds reseeding by 15.7\,pp (95\% CI 10.5 to 20.8); among pairs matched at $r > 0.9$ its effects are as close as a reseed's (median difference $+0.014$, CI $-0.018$ to $+0.042$). Seed 0, one sample.}
\label{tab:featablation}
\centering\small
\begin{tabular}{lrrrrrr}
\toprule
condition & $r > 0.9$ & same effect & same $|\,r > 0.9$ & effect cos & $\Delta$logp $r$ & top-50 ctx \\
\midrule
identical bytes & 100.0 & 100.0 & 100.0 & 1.000 & 1.000 & 1.00 \\
1 ulp on $10^{-6}$ of values & 73.0 & 79.7 & 94.5 & 0.971 & 0.966 & 0.82 \\
fp16 & 63.7 & 68.0 & 89.5 & 0.942 & 0.935 & 0.78 \\
std-int4 & 38.7 & 38.7 & 73.3 & 0.842 & 0.816 & 0.62 \\
na\"ive int4 & 15.7 & 12.7 & 46.8 & 0.651 & 0.584 & 0.38 \\
reseed, seed 1 & 26.0 & 22.7 & 53.8 & 0.748 & 0.687 & 0.48 \\
reseed, seed 2 & 22.3 & 23.3 & 64.2 & 0.727 & 0.680 & 0.46 \\
\bottomrule
\end{tabular}
\end{table}

\begin{table}[htbp]
\caption{Is a lost feature gone or re-cut? Share of alive fp32 features recovered on held-out tokens by a least-squares map from their $m$ best-correlated partners ($R^2 > 0.81$, so $m{=}1$ is the retraining count re-scored on these tokens). Recovering from four partners keeps std-int4 13.4\,pp above reseeding (three seeds each). Lost: of the features lost at $m{=}1$, the share four partners recover, 12--15\% in every condition but na\"ive int4; four random partners recover at most 0.12\%. Set EV: the lost features' top-64 decoder subspace, applied to the retrained SAE's unmatched features, as a fraction of its within-SAE value~\citep{reprosubspaces}. $\pm$ rows are three runs.}
\label{tab:subspace}
\centering\small
\begin{tabular}{lrrrrr}
\toprule
condition & $m{=}1$ & $m{=}4$ & $m{=}8$ & lost, $m{=}4$ & set EV \\
\midrule
identical bytes & 100.0 & 100.0 & 100.0 & --- & --- \\
1 ulp on $10^{-6}$ of values & $70.4 \pm 1.1$ & $74.0 \pm 1.2$ & $75.2 \pm 1.1$ & $13.2 \pm 1.2$ & $0.89 \pm 0.00$ \\
fp16 & 60.1 & 65.6 & 67.3 & 14.0 & 0.91 \\
rot-int4 & 37.7 & 46.7 & 50.0 & 14.8 & 0.93 \\
std-int4 & $37.9 \pm 0.8$ & $46.6 \pm 0.8$ & $49.7 \pm 0.8$ & $14.3 \pm 0.1$ & $0.91 \pm 0.01$ \\
na\"ive int4 & 15.6 & 22.4 & 26.2 & 8.5 & 0.73 \\
reseed & $22.9 \pm 0.3$ & $33.2 \pm 0.4$ & $37.8 \pm 0.4$ & $13.5 \pm 0.4$ & $0.92 \pm 0.00$ \\
\bottomrule
\end{tabular}
\end{table}

\begin{table}[htbp]
\caption{Fresh text against replay at the $3\times$ budget (98.3M token-visits). Floors are three seed pairs; codec rows are same-seed pairs at three seeds, all $\pm$ half the range. Stable: features matched above $r = 0.9$ in both other fp32 seeds of the same regime; fixed-774: the top 774 by cross-seed agreement, as in Table~\ref{tab:budget}. The fresh floor sits +0.09\,pp from the replay floor and std-int4 keeps the same lead in both regimes.}
\label{tab:fresh}
\centering\small
\begin{tabular}{lrrrrr}
\toprule
regime & SAE & agreement \% & $\times$floor & stable \% & fixed-774 \% \\
\midrule
replay (20M, $\sim$5 passes) & floor (fp32 reseed) & $33.3 \pm 0.1$ & 1.00 & --- & --- \\
 & fp16 & $49.0 \pm 0.7$ & 1.47 & $92.0 \pm 0.2$ & $95.8 \pm 0.4$ \\
 & std-int4 & $38.7 \pm 0.6$ & 1.16 & $87.8 \pm 1.9$ & $93.7 \pm 1.7$ \\
\midrule
fresh (98.5M, one pass) & floor (fp32 reseed) & $33.4 \pm 0.3$ & 1.00 & --- & --- \\
 & std-int4 & $38.8 \pm 0.3$ & 1.16 & $88.0 \pm 1.5$ & $93.6 \pm 1.2$ \\
\bottomrule
\end{tabular}
\end{table}

\begin{table}[htbp]
\caption{Equal stored bytes at $3\times$: fp16 on the 20M-token cache against std-int4 on the first 79.0M tokens of the fresh store, in 99.76\% of fp16's bytes. Agreement is cross-seed against fresh fp32 (six pairs per candidate, mean and range), so no candidate shares an initialization or data order with its reference; the reference row is fresh fp32 against itself across seeds (three pairs). The two candidates differ by +0.02\,pp: a tie. $^\dagger$Post-hoc resolution check, not pre-registered (na\"ive int4: seed 0 only, two pairs).}
\label{tab:equalbytes}
\centering\small
\begin{tabular}{lrrrr}
\toprule
candidate & tokens (M) & stored GB & agreement \% & $\times$ref \\
\midrule
fp32, fresh (reference) & 98.5 & 303 & 33.4 (33.2--33.7) & 1.00 \\
fp16, replayed & 20.0 & 30.7 & 32.6 (32.0--33.0) & 0.974 \\
std-int4, more text & 79.0 & 30.6 & 32.6 (31.4--33.3) & 0.974 \\
\midrule
fp32, replayed$^\dagger$ & 20.0 & 61.4 & 32.8 (31.8--33.2) & 0.980 \\
na\"ive int4, replayed$^\dagger$ & 20.0 & 7.8 & 15.9 (15.7--16.1) & 0.476 \\
\bottomrule
\end{tabular}
\end{table}

\begin{table}[htbp]
\caption{Training throughput on an A100 (thousand tokens/s, full $1\times$ run, mean of two fresh containers) by cache format and where the loader decodes; busy is the share of training time the loader spent reading, decoding and shuffling. Every decode path of a format trains bit-identical weights. Best std-int4 over best fp32: 1.09 from local disk, 1.04 from the network volume, under the pre-registered $1.10$ for a material gain. Volume reads varied 60--840\,MB/s between runs.}
\label{tab:timing}
\centering\small
\begin{tabular}{lrrrrr}
\toprule
format & decode & local & busy & volume & busy \\
\midrule
fp32 & CPU (paper) & 117.0 & 0.83 & 75.1 & 0.99 \\
 & GPU & 118.5 & 0.51 & 119.5 & 0.53 \\
fp16 & CPU (paper) & 114.6 & 0.89 & 115.1 & 0.84 \\
 & GPU & 121.0 & 0.28 & 119.7 & 0.33 \\
std-int4 & CPU (paper) & 92.9 & 0.99 & 50.1 & 1.00 \\
 & CPU, no matmul & 92.6 & 0.99 & 90.9 & 0.99 \\
 & GPU & 129.1 & 0.08 & 123.8 & 0.20 \\
\bottomrule
\end{tabular}
\end{table}

\begin{table}[htbp]
\caption{The frozen-SAE test explained by TopK margins (GPT-2 layer 6, the frozen test's own SAE and tokens). model: Spearman correlation, over alive features, between predicted and measured frozen disagreement $1-r$; the prediction uses only the second moment of the codec's error, estimated on held-out tokens, and the SAE's margins on clean activations. statistic: the largest $|\rho|$ of log firing rate, log mean or log peak activation with $1-r$. crossing \%: median share of a degraded or damaged feature's squared error that comes from tokens where it is active on one side only (--: not reported for the noise-like codecs, which have at most 6 degraded or damaged features each). Below the rule, codecs that remove part of the signal rather than add noise. Sources: \texttt{geom\_margins.json}, \texttt{geom\_summary.json}.}
\label{tab:margins}
\centering\small
\begin{tabular}{lrrrr}
\toprule
codec & frozen mean $r$ & model $\rho$ & statistic $|\rho|$ & crossing \% \\
\midrule
fp16 & 1.0000 & 0.736 & 0.644 & -- \\
bf16 & 0.9998 & 0.888 & 0.734 & -- \\
int8 & 0.9963 & 0.989 & 0.843 & -- \\
int8-o8 & 0.9987 & 0.966 & 0.818 & -- \\
int6 & 0.9832 & 0.991 & 0.861 & -- \\
int6-o8 & 0.9948 & 0.979 & 0.838 & -- \\
rot-int8 & 0.9992 & 0.959 & 0.795 & -- \\
rot-int6 & 0.9969 & 0.984 & 0.818 & -- \\
rot-int4 & 0.9854 & 0.988 & 0.830 & -- \\
std-int4 & 0.9847 & 0.980 & 0.835 & -- \\
int4-o16 & 0.9821 & 0.992 & 0.869 & -- \\
\midrule
int4 & 0.8401 & 0.861 & 0.875 & 84 \\
pca256 & 0.8098 & 0.877 & 0.829 & 46 \\
rp256 & 0.5796 & 0.608 & 0.681 & 57 \\
\bottomrule
\end{tabular}
\end{table}

\begin{table}[htbp]
\caption{Predicting from one SAE which features retraining recovers (GPT-2 layer 6, AUC). Above the rule the target is seed stability (best match $r>0.9$ in both other seeds); the logistic model is fit once on the $1\times$, $k=32$ SAE (its row is cross-validated) and applied unchanged to the other regimes' own SAEs. Below the rule, the same seed-fit model predicts which features survive retraining on each codec. positive \%: seed-stable or retained share of alive features. statistics: log firing rate, log mean and log peak activation. geometry: co-firing share with the best partner, decoder coherence, co-firing of the nearest decoder atom, TopK margin relative to the activation, encoder-decoder cosine. Predictors use held-out eval tokens. Source: \texttt{geom\_identifiability.json}.}
\label{tab:stabpredict}
\centering\small
\begin{tabular}{lrrrr}
\toprule
target & positive \% & statistics & geometry & both \\
\midrule
$1\times$, $k=32$ (fit) & 15.9 & 0.846 & 0.829 & 0.868 \\
$3\times$ & 24.9 & 0.787 & 0.741 & 0.794 \\
$10\times$ & 32.9 & 0.762 & 0.696 & 0.752 \\
$3\times$ fresh text & 24.9 & 0.793 & 0.738 & 0.794 \\
$k=16$ & 21.0 & 0.817 & 0.801 & 0.837 \\
$k=64$ & 10.1 & 0.899 & 0.842 & 0.902 \\
\midrule
fp16 survival & 60.6 & 0.832 & 0.743 & 0.808 \\
rot-int4 survival & 37.7 & 0.841 & 0.760 & 0.828 \\
std-int4 survival & 38.0 & 0.830 & 0.762 & 0.824 \\
\bottomrule
\end{tabular}
\end{table}

\begin{table}[htbp]
\caption{In this run, when a perturbation reaches training decides its cost (GPT-2 layer 6, seed 0, one run per row). Each perturbed run is a copy of the reference run, which reproduces Table~\ref{tab:m1full}'s fp32 seed-0 SAE bit for bit, that receives the change at the listed step(s) and trains on the same batches. agreement: the paper's retained percentage against the reference. distance: relative Frobenius distance of the decoder from the reference's, at the end and 2,000 steps after the change. The one-ulp row replays the one-scalar experiment (its altered value enters training twice; the weights are bit-identical to the reference at every logged step from 130 until the second entry; optimizer moments were not compared) and reproduces that SAE bit for bit. The last row is not a copy: it is the separately trained noise7 SAE of Table~\ref{tab:dose}, where every batch carries the change, scored against the same reference. Sources: \texttt{geom\_summary.json}, \texttt{pair\_geomdyn\_*.json}.}
\label{tab:dynamics}
\centering\small
\begin{tabular}{lrrrr}
\toprule
change & step & agreement \% & distance & +2{,}000 steps \\
\midrule
$10^{-7}$, one batch & 0 & 64.7 & 0.6561 & 0.2338 \\
 & 4{,}000 & 99.9 & 0.0830 & 0.0452 \\
 & 6{,}000 & 100.0 & 0.0301 & 0.0301 \\
 & 7{,}000 & 100.0 & 0.0198 & -- \\
 & 7{,}500 & 100.0 & 0.0107 & -- \\
 & 7{,}800 & 100.0 & 0.0041 & -- \\
 & 7{,}950 & 100.0 & 0.0001 & -- \\
\midrule
one ulp, one value & 112, 4{,}970 & 99.98 & 0.0562 & -- \\
$10^{-7}$, every batch & all & 66.0 & 0.6447 & -- \\
\bottomrule
\end{tabular}
\end{table}